\documentclass[10pt,twocolumn,letterpaper]{article}

\usepackage{cvpr}[preprint]
\usepackage{multirow}
\usepackage{fontawesome5}
\usepackage{tabularx}
\usepackage{titletoc}

\newcommand{\datasetName}{AssemblyBench\xspace}

\newcommand{\modelName}{AssemblyDyno\xspace}

\DeclareMathOperator{\SE}{SE}
\newcommand{\order}{\ensuremath{\pi}}
\newcommand{\permuteMat}{M}

\renewcommand{\subsubsection}[1]{\noindent\textbf{#1}}

\usepackage{subcaption}

\usepackage{listings}
\usepackage{xcolor}

\definecolor{mdgray}{gray}{0.95}
\definecolor{mdbrown}{RGB}{150,100,0}
\definecolor{mdblue}{RGB}{0,102,204}
\definecolor{mdgreen}{RGB}{0,128,0}
\definecolor{mdpurple}{RGB}{150,0,150}

\lstdefinelanguage{Markdown}{
    basicstyle=\rmfamily,
    sensitive=false,
    comment=[l]{\#},                 
    moredelim=[is][\bfseries]{**}{**},      
    moredelim=[is][\itshape]{*}{*},          
    moredelim=[is][\ttfamily]{`}{`},         
}

\lstdefinestyle{markdownStyle}{
    backgroundcolor=\color{mdgray},
    basicstyle=\rmfamily \footnotesize,
    frame=single,
    breaklines=true,
    columns=fullflexible,
    keepspaces=true
}

\usepackage[accsupp]{axessibility}

\definecolor{cvprblue}{rgb}{0.21,0.49,0.74}
\usepackage[pagebackref,breaklinks,colorlinks,allcolors=cvprblue]{hyperref}
\usepackage{xspace}

\title{AssemblyBench: Physics-Aware Assembly of Complex Industrial Objects}

\vspace{-2mm}
\author{
Danrui Li$^1$\thanks{Work done during internships at MERL.}
\qquad Jiahao Zhang$^{2*}$
\qquad Bernhard Egger$^3$ \\
Moitreya Chatterjee$^4$
\qquad Suhas Lohit$^4$
\qquad Tim K. Marks$^4$ 
\qquad Anoop Cherian$^4$
\\
\small{$^1$Rutgers, The State University of New Jersey, USA
\qquad$^2$The Australian National University, Australia}\\
\small{\qquad$^3$Friedrich-Alexander-Universit\"{a}t Erlangen-N\"{u}rnberg, Germany
$^4$Mitsubishi Electric Research Laboratories (MERL), USA}\\
\scriptsize{
$^1${\tt danrui.li@rutgers.edu}
\quad$^2${\tt jiahao.zhang@anu.edu.au}
\quad$^3${\tt bernhard.egger@fau.de}
\quad$^4${\tt\{chatterjee,slohit,tmarks,cherian\}@merl.com}
}\\
\begingroup
\hypersetup{urlcolor=red}
\small{\url{https://merl.com/research/highlights/assemblybench}}
\endgroup
}

\begin{document}
\maketitle

\begin{abstract}
Assembling objects from parts requires understanding multimodal instructions, linking them to 3D components, and predicting physically plausible 6-DoF motions for each assembly step. Existing datasets focus on simplified scenarios, overlooking shape complexities and assembly trajectories in industrial assemblies. We introduce \datasetName, a synthetic dataset of 2,789 industrial objects with multimodal instruction manuals, corresponding 3D part models, and part assembly trajectories. We also propose a transformer-based model, \modelName, which uses the instructional manual and the 3D shape of each part to jointly predict assembly order and part assembly trajectories. \modelName outperforms prior works in both assembly pose estimation and trajectory feasibility, where the latter is evaluated by our physics-based simulations.
\end{abstract}

\section{Introduction}
\label{sec:intro}

\begin{figure}[htbp]
    \centering
    \includegraphics[width=.9\linewidth]{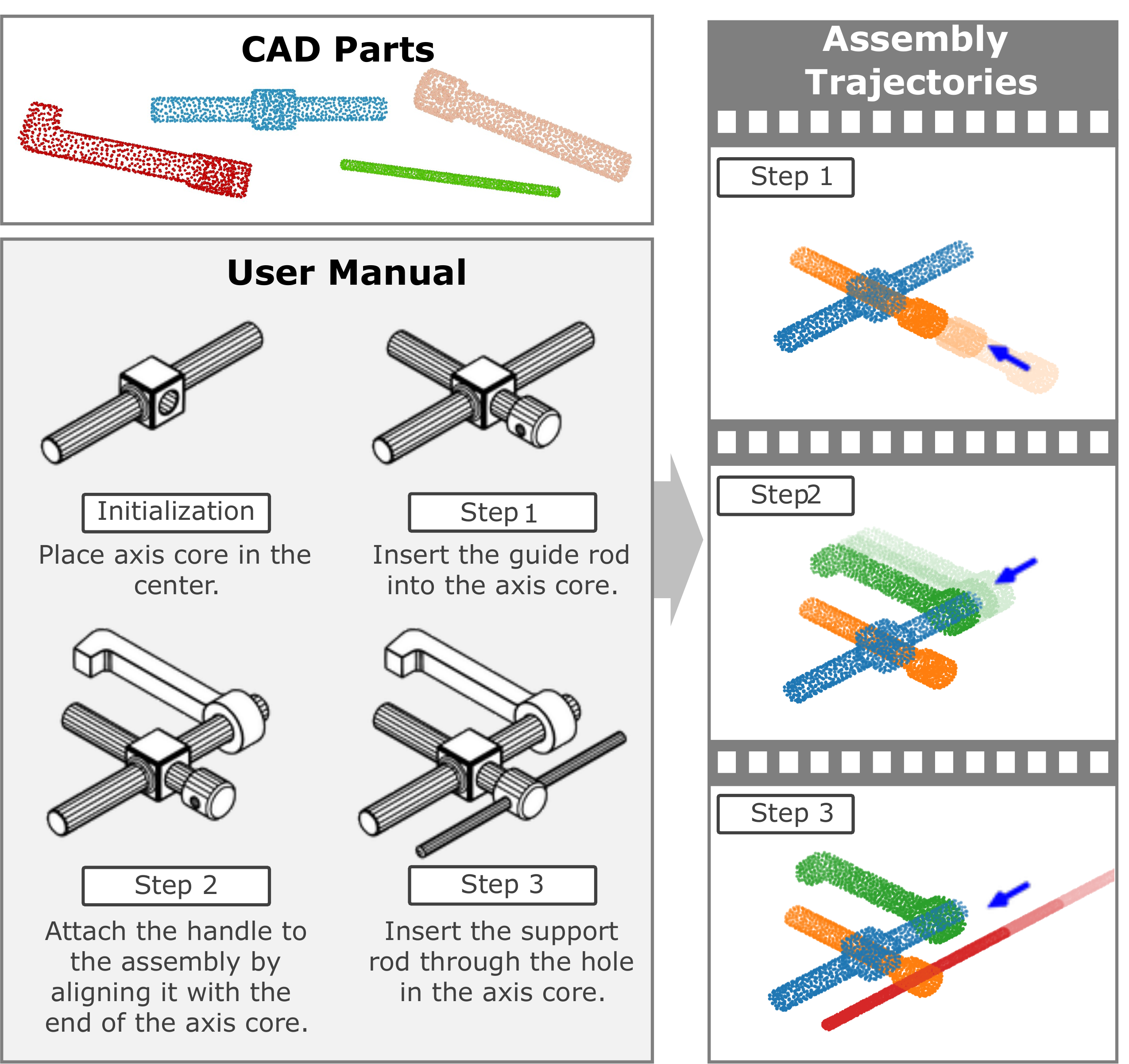}
    \vspace{-6pt}
    \caption{Given a step-wise manual with diagrams and text (\emph{lower left}), we aim to assemble the corresponding set of 3D parts (\emph{upper left}) in a virtual environment, outputting its step-wise assembly trajectories, which can be rendered into 4D animations (\emph{right}). }
    \label{fig:teaser}
    \vspace{-3mm}
\end{figure}

Assembling objects from constituent parts is a challenging yet ubiquitous task with substantial potential for automation, and it has myriad applications from household furniture assembly to large-scale manufacturing of complex industrial objects. As a result of the advancements in large vision-and-language models and robotics foundation models, the problem of object assembly has garnered significant interest recently in both the computer vision and robotics communities~\cite{li2020imagepa,zhang2025manual,tie2025manual2skill,tian2025fabrica,tang2024automate}. State-of-the-art approaches to address this task mainly consider IKEA-style furniture assembly, due to the availability of abundant collections of well-designed instruction manuals that detail each step of the process using language-free diagrams~\cite{wang2022ikea,zhang2025manual,tie2025manual2skill}. Furthermore, to facilitate assembly by inexperienced users, furniture parts are typically designed to be easily distinguishable, clearly illustrated in diagrams, and straightforward to attach to each other. Thus, datasets derived from such furniture assemblies offer a simplified yet useful setup to study this complex reasoning task.

However, furniture assemblies alone may not capture the full spectrum of complexities in real-world assembly processes, especially the process of moving assembly parts. For example, the assembly of electrical appliances (\emph{e.g.}, air conditioners, ceiling fans, laundry machines), industrial equipment (\emph{e.g.}, motors, gear boxes, hydraulic pumps), or even simple interactive toys. These objects often contain parts with complex geometries, and they may require sophisticated maneuvers such as insertion with twisting to assemble.  While there have been several attempts at capturing varied aspects of this complex problem---\emph{e.g.}, assembly from diagram-based instruction manuals~\cite{zhang2025manual,tie2025manual2skill,tie2025manual2skillconnect}, planning for robotic assembly~\cite{tang2024automate,tian2025fabrica}, and learning from video demonstrations~\cite{Zhang2023Aligning,ben2021ikea,liu2024ikea}---there is a need for datasets that capture more of the common challenges in assembly.

Toward this end, we present \datasetName, a novel synthetic assembly dataset. It consists of nearly 3K assemblies spanning several categories of objects (not merely furniture) and featuring industrial objects.
In contrast to prior non-IKEA datasets~\cite{tian2024asap,zhang2025manual}, \datasetName extends the dataset modalities from assembled shapes to a set of CAD parts, step-by-step instruction diagrams with text descriptions, and assembly motion trajectories.
Moreover, we introduce a pipeline that can be generalized for automatic instruction manual creation for any type of industrial objects from their CAD assemblies, which are commonly provided in mechanical design specifications.

    To address the challenges in \datasetName,  we present \emph{\modelName}, a novel transformer-based architecture that predicts the assembly order of the parts as well as their 6-DoF motion trajectories. Specifically, \modelName is trained to learn a soft attention between the order of the instructions and the 3D part point clouds (encoded using a point cloud encoder) to regressing a discrete sequence of $SE(3)$ transformations for each part, along which the part should move in order to successfully complete its assembly. The entire set of motion sequences is jointly predicted by \modelName in a single forward pass. Our model is trained in a supervised setting, using ground-truth sequences of part trajectories, by minimizing chamfer-distance-based losses that account for invariance to symmetries in the parts' geometries.

    In addition to utilizing the standard metrics for evaluating the performance of prior works on \datasetName (\emph{e.g.}, symmetric chamfer distance and final success rate), we present a novel evaluation that executes the predicted part motion trajectories in a physics simulator~\cite{newton2025} to verify their physical feasibility. Our key insights are three fold: i) while our training scheme does not include the simulator-in-the-loop, our supervised training might implicitly capture the physical constraints for assembly; ii) although instruction manuals are usually created to follow physically plausible part assembly, there may be other, novel motion pathways that could lead to a correct assembly; and iii) there may be inaccurate or physically infeasible steps in the predicted assembly that cannot be identified unless executed under physical constraints. For example, a predicted motion may cause the part to get stuck in an intermediate position, which would prevent the remainder of the assembly from proceeding. While prior protocols measure success using only the point-cloud alignment of the predicted final assembly, our physics-simulator-based evaluation offers a complete verification of the part assembly order, motion trajectories, and physical realizability, bringing successful assemblies significantly closer to real-world enactment. 

We present extensive experiments demonstrating various aspects of \datasetName using \modelName. We find that a state-of-the-art baseline~\cite{zhang2025manual}, which demonstrates nearly 60\% success rates on furniture datasets, does not perform nearly as well (nearly 30\% worse) on our challenging new dataset. In contrast, \modelName, with its incorporation of both diagrams and text descriptions from the instruction manual, leads to 12\% improvements in the success rate of final pose estimate. Furthermore, \modelName predicts the assembly trajectories with better physics feasibility. It achieves about 33\% success rate in a physical simulator by referencing to diagrams and texts with a trajectory smoothing loss, while the baseline method achieves only around 3\%.

In summary, our main contributions include:
\begin{itemize}
\item \textbf{Dataset:} \datasetName that includes complex industrial part assemblies with multi-modal user manuals and assembly trajectories, produced using a VLM-based generative pipeline. 
\item \textbf{Model:} \modelName, a generalized feed-forward model that takes in multi-modal assembly steps and 3D part point clouds, predicting the parts' assembly order, final poses, and 6 DoF assembly motion trajectories.
\item \textbf{Evaluation:} A physics engine based protocol to evaluate the physical feasibility of predicted assembly trajectories, where \modelName shows state-of-the-art results.
\end{itemize}

\section{Related Work}
\label{sec:related}
\subsubsection{Assembly Datasets.}
A comprehensive assembly dataset should include diverse geometries and contact types, incorporate realistic physical interactions, be realizable in a physics simulator, and capture the full assembly process rather than only final poses. Existing shape datasets such as PartNet~\cite{Mo_2019_Partnet} and IKEA-based assembly datasets~\cite{wang2022ikea,zhang2025manual,tie2025manual2skill,tie2025manual2skillconnect,liu2024ikea,ben2021ikea,Zhang2023Aligning} have been widely used, with IKEA manuals providing canonical step-by-step diagrammatic supervision. However, these datasets include a limited set of furniture objects with similar part geometries and lack kinematic constraints. Broader datasets covering toys~\cite{Sener_2022_Assembly101} or electronics~\cite{sliwowski2025reassemble}, as well as datasets with real-world video annotations~\cite{Sener_2022_Assembly101,EPIC-Tent,zheng2023havid,sliwowski2025reassemble}, broaden category coverage but still focus mainly on high-level goals or final part poses.

Motivated by these limitations and following~\cite{tian2024asap}, we build on the Assemble-Them-All (ATA) dataset~\cite{tian2022assemble,tian2024asap}, which contains nearly 5K industrial CAD models with explicit part-insertion relations and physics-based disassembly trajectories~\cite{tian2022assemble}. Prior work has used ATA~\cite{tian2024asap,zhu2023multilevelassembly}, but it lacks step-by-step manuals, standardized part/trajectory representations, curated splits, and evaluation protocols.

Recent efforts aim to reduce manual annotation cost via automatic manual-generation pipelines, including parametric systems~\cite{patel2025dynamo,lego2plan,Pun_2025_BrickGPT} and VLM-based dataset enrichment~\cite{text2cad,xu2025cadmllm,cadllama2025}. CheckManual~\cite{checkmanual} further explores VLM-driven operation-manual generation. However, assembly involves multi-part interactions, occlusions, and nontrivial 3D insertions, making manual generation substantially more challenging. Consequently, most datasets~\cite{zhang2025manual,hasegawa-etal-2025-promqa-assembly,wang2022ikea} provide only final part poses, overlooking the trajectories required for complex assemblies. Our dataset and pipeline address these gaps by producing standardized representations, full assembly trajectories, and step-by-step manuals. See Table~\ref{tab:dataset-compare} for detailed comparison.

\begin{table*}[htbp]
    \footnotesize
    \centering
    \caption{
    Comparison of assembly/manipulation datasets. 
    Legends: \faImages[regular] diagrams, \faList*[regular] text instructions, \faFilm videos, \faDiceD6 3D objects.
    }
    \vspace{-6pt}
    \begin{tabular}{lcccc}
    \toprule
         Work & Domain & Input & Output & Size \\
    \midrule
    IKEA-Manual~\cite{wang2022ikea} 
    & Furniture 
    & \faImages[regular], \faDiceD6 
    & Assembly order, 6D part pose
    & 102 \\

    IKEA-Manuals-at-Work~\cite{liu2024ikea} 
    & Furniture  
    & \faImages[regular], \faFilm, \faDiceD6 
    & 6D part pose, temporal alignments
    & 36 \\

    ProMQA‑Assembly~\cite{hasegawa-etal-2025-promqa-assembly}
    & Toys
    & \faImages[regular], \faList*[regular], \faDiceD6 
    & Assembly order, General QA
    & 78 \\

    LEGO‑VLM~\cite{huang2025legocobuild}
    & Modular toy bricks
    & \faImages[regular], \faList*[regular], \faDiceD6 
    & Object localization, state judgement
    & 65 \\

    MEPNet~\cite{lego2plan}
    & Modular toy bricks
    & \faImages[regular], \faDiceD6 
    & Assembly order, part pose
    & 8000 \\

    ManualPA~\cite{zhang2025manual} 
    & Furniture 
    & \faImages[regular], \faDiceD6 
    & Assembly order, 6D part pose 
    & 6871 \\

    CheckManual~\cite{checkmanual} 
    & Appliances 
    & \faImages[regular], \faList*[regular], \faDiceD6 
    & Manipulation plan, action trajectory 
    & 369 \\

    Manual2Skill++~\cite{tie2025manual2skillconnect} 
    & Furniture, toys, industrial 
    & \faImages[regular], \faDiceD6 
    & Assembly order, 6D pose trajectories
    & 20+ \\

    \textbf{\datasetName (Ours)} 
    & All the above except toy bricks
    & \faImages[regular], \faList*[regular], \faDiceD6 
    & Assembly order, 6D pose trajectories
    & 2789 \\
    \bottomrule
    \end{tabular}

    \label{tab:dataset-compare}
\end{table*}

\subsubsection{Assembly Step Prediction.} There are numerous works that attempt to predict assemblies without manuals~\cite{li2024joint,pmlr-v229-li23a,spaformer,compoNet}. However, given the joint discrete-and-continuous search space for finding the part to assemble and generating its assembly trajectory, it is usually difficult for such methods to generalize. There are are also several recent works that use guidance from: i) final object renderings~\cite{li2020imagepa,zhu2023multilevelassembly,Wu_2020_pqnet}, ii) from step-wise manuals~\cite{zhang2025manual,tie2025manual2skill,tie2025manual2skillconnect}, or iii) assembly videos~\cite{liu2024ikea,Zhang2023Aligning}, but mostly for IKEA-type furniture. Classical motion planning methods (e.g., RRT~\cite{LaValle1998RRT} and PRM~\cite{kavraki1996}) have been explored for assembly via generating the motion trajectory from a predicted final part pose~\cite{zhu2023multilevelassembly}, including incorporating physical constraints~\cite{tian2022assemble,tian2024asap}. However, they are computationally expensive and require precise characterization of the physical constraints in the environment~\cite{kingston2018review}. Thus, while we use a physics-based planner~\cite{tian2022assemble,tian2024asap} when generating our dataset, our model is trained in a supervised manner on the trajectories, implicitly learning the physics. Further, our use of a single forward pass to predict all assembly steps at once in discrete time steps is computationally efficient and robust.


\section{Proposed Method}
\label{sec:method}
In its generalized form, an assembly task involves understanding the procedure from instruction manuals, identifying the object parts to be assembled at each step, and executing the assembly steps as depicted in the manual to fit the parts together following physically feasible motion paths. Following this recipe, we formulate our task as follows.

We are given an unordered set of $N$ assembly parts as 3D point clouds $\{P_i\}_{i=1}^N$, with each part assumed to contain the same number of points), and a manual consisting of a sequence of assembly instructions denoted $(\mathcal{I}_1,\cdots, \mathcal{I}_N)$, where each step involves adding one part. Our objective is to have a model that: i) predicts the assembly order by grounding the 3D parts to their instructions, i.e., producing part indices $(\hat{\order}_1, \hat{\order}_2, \cdots, \hat{\order}_N)$ such that the part $\mathcal{P}_{\hat{\order}_i}$ is associated with instruction $\mathcal{I}_i$,
and ii) predicting the part motion trajectories for each assembly step as 
$\bigl((\hat{R}_i^k, \hat{t}_i^k)\in \SE(3)\bigr),$
where $i \in \{1, \ldots, N\} $ represents the assembly step number, and $k \in \{1, \ldots, T\}$ represents the time step within a part's trajectory.
The number of parts $N$ is assumed to vary across objects, however the number of time steps $T$ in each part's assembly trajectory is considered fixed.
Each instruction step $\mathcal{I}_i$ in the manual consists of a diagram illustrating the assembly step and a free-form text description detailing the step. We use 2D line-drawing diagrams similar to IKEA manuals, showcasing 3D parts without textures in a fixed parallel projection. The assembly trajectory in assembly step $i$ is represented as a sequence of $T$ 6-DoF poses, $\bigl(\hat{R}_i^k, \hat{t}_i^k \bigr)$, which respectively represent the $3 \times 3$ rotation matrix and $3 \times 1$ translation vector of the part at the $k$th time-step of its trajectory. In our experiments, we use $T=12$ for the number of time-steps. 

This work requires us to implement three parts: i) building the \datasetName dataset using an automatic annotation pipeline, which adheres to the assembly process described above while incorporating complex assemblies (detailed in \S~\ref{sec:dataset-pip}); ii) proposing a novel transformer-based reasoning model, \modelName, which predicts an entire 3D assembly process from an instruction manuals (described in \S~\ref{sec:model}); and iii) proposing a set of evaluation metrics that evaluates the entirety of the assembly performance (explained in \S~\ref{sec:eval}). 

\subsection{\datasetName Dataset}
\label{sec:dataset}

\begin{figure*}
    \centering
    \includegraphics[width=.85\linewidth]{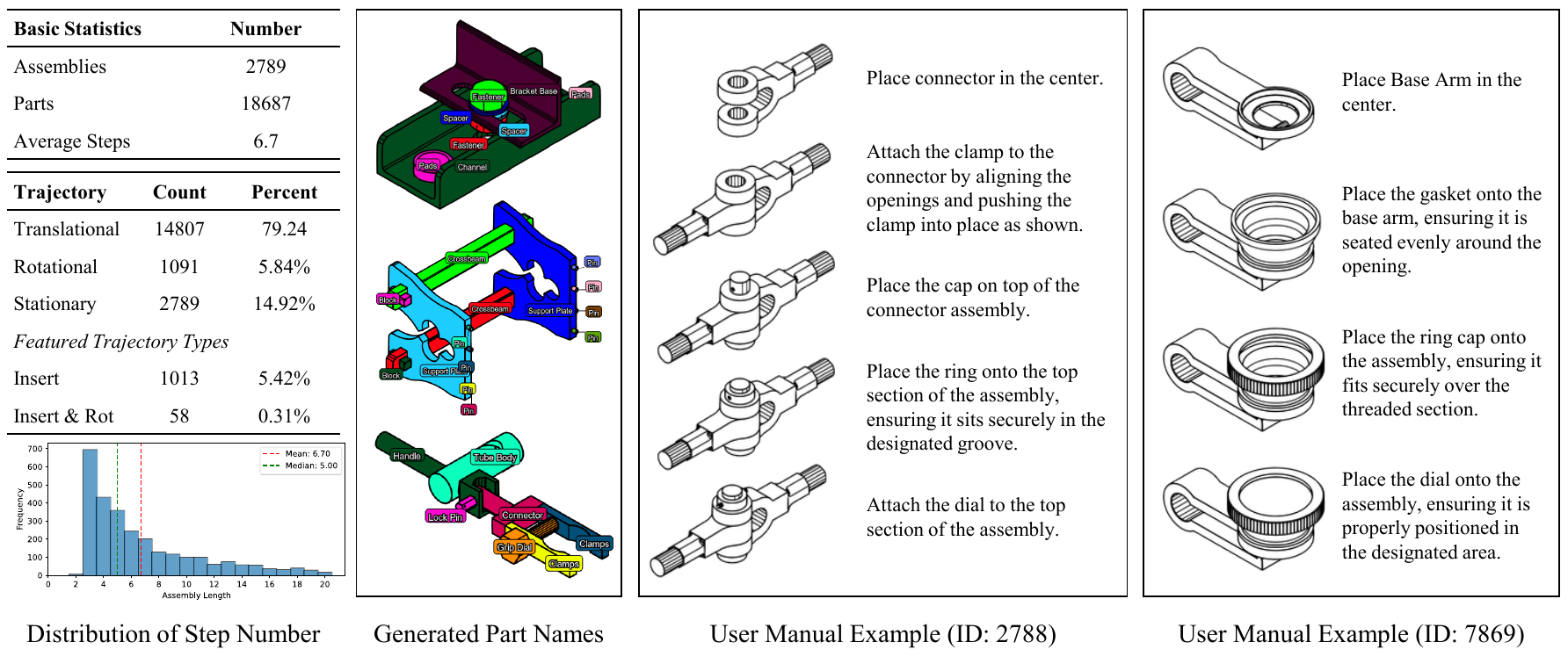}
    \vspace{-6pt}
    \caption{\textbf{Overview of \datasetName}. \emph{Column 1:} Statistics of our \datasetName dataset and histogram of the number of parts per assembly in \datasetName. \emph{Column 2:} Generated part names in \datasetName. The coloring and the labels are for visualization in this figure only---they are not included in the model inputs. \emph{Columns 3--4:} Example generated instruction manuals for two different assemblies.}
    \label{tab:data-stats}
    \label{fig:manual}
    \label{fig:step-hist}
    \label{fig:part-name}
    \vspace{-3mm}
\end{figure*}

\datasetName contains multimodal instruction manuals for 2789 assemblies in total, covering a wide range of categories including furniture, appliances, and mechanical components. Each manual provides the 3D mesh and point cloud of each individual part, as well as step-wise assembly instructions consisting of diagrams and text.
The number of steps (i.e., the number of parts) for each assembly ranges from 2 to 20, with an average of 6.7 steps on average.
Table~\ref{tab:dataset-compare} contrasts \datasetName with prior datasets proposed for assembly-related tasks. As is clear, \datasetName generalizes prior works while including ground-truth part assembly trajectories, exhibiting a variety of motion patterns. In Figure~\ref{tab:data-stats}, we provide detailed statistics on the properties of our dataset. Out of all of the trajectories, 5.84\% involve rotational movements such as sophisticated twists, 5.42\% involve long translation movements that indicate insertions into a hole or slot, and there are about 58 parts that need long distance insertions combined with rotations for the assembly. We divide the set of 2789 assemblies into train/val/test splits in 80\%-10\%-10\% allocation.

\subsection{\datasetName Construction Pipeline} 
\label{sec:dataset-pip}

\begin{figure*}
    \centering
    \includegraphics[width=.8\linewidth]{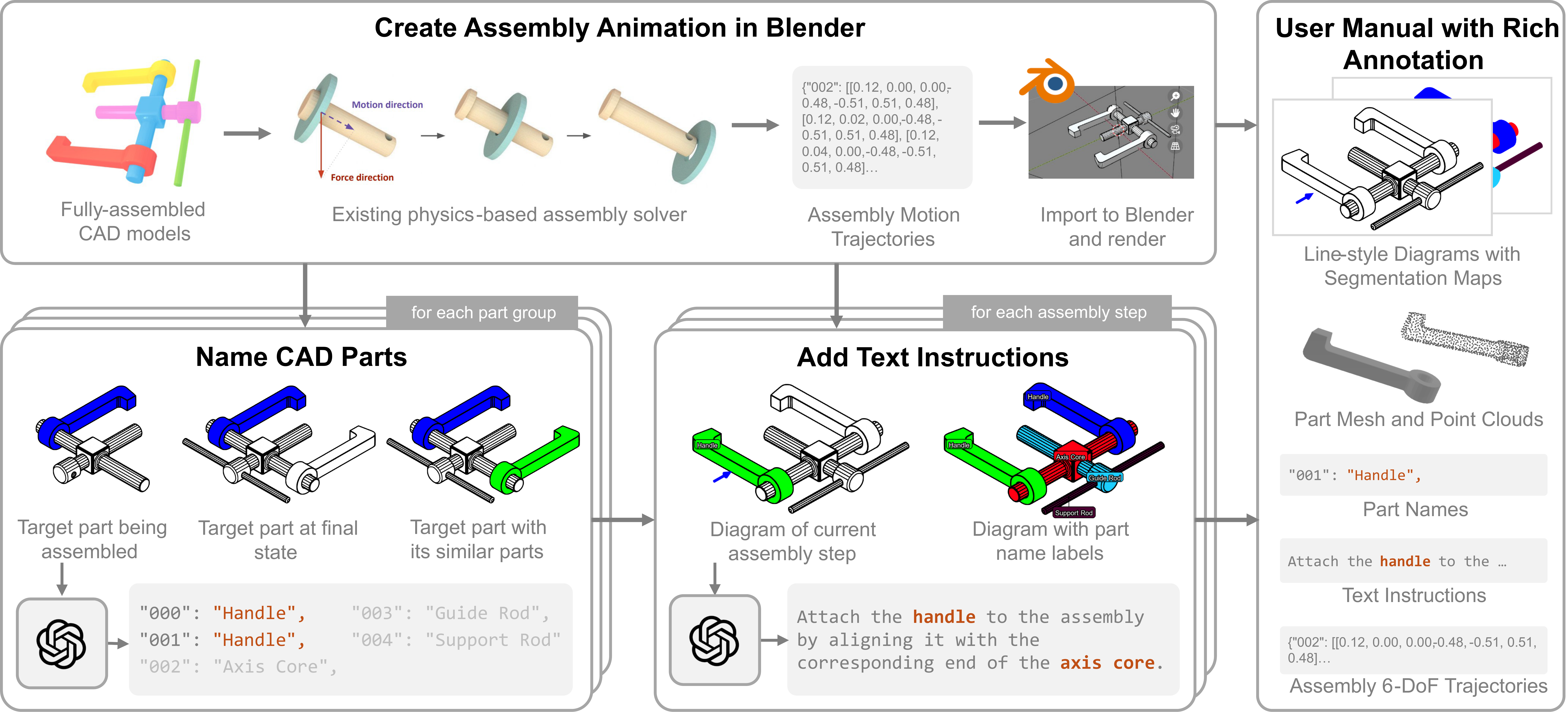}
    \vspace{-6pt}
    \caption{\textbf{Manual creation pipeline for \datasetName}. 
    \emph{Top left:} From a CAD model of an assembled object, we calculate the part assembly trajectories using a physical engine and import the animations to Blender. \emph{Bottom left:} Blender renderings are fed to VLMs to create CAD part names and textual assembly instructions. \emph{Right:} All annotations are used to generate a single step in the final manual. 
    }
    \label{fig:dataset}
    \vspace{-3mm}
\end{figure*}

 The following subsections present our automatic data generation pipeline of \datasetName, illustrated in Figure~\ref{fig:dataset}. We note that our pipeline is very general and could be used to generate assembly instruction manuals for a broad variety of objects, given only an object's 3D CAD model. Such CAD models of assembled real-world objects are widely available (e.g., from machine designs).

\subsubsection{Part Order and Motion Trajectories:}
\label{sec:anim-gen}
There are two important sub-tasks in assembly: i) deciding the order of the parts to select for the assembly (so that it is physically realizable) and ii) planning the motion of each part from its initial pose to the final assembled pose. The assembly-by-disassembly process~\cite{tian2024asap,tian2022assemble} tackles these sub-tasks via importing the 3D CAD models into a physics engine. Specifically, it uses a depth-first-search algorithm to attempt to disassemble the object one part at a time via applying forces along the part axes, until the part becomes disassociated from the other parts. This scheme discovers a disassembly sequence that includes both a disassembly order of parts and a 6-DoF pose trajectory for removing each part. Reversing both of these yields both the assembly part order and the assembly pose trajectories. To create \datasetName, we import each CAD object's assembly steps and motion trajectories (discretized to $T$ time steps) to Blender~\cite{hess2007essential} and format them to produce assembly animations.

\subsubsection{Diagram Generation:}
\label{sec:diagram-gen}
A key ingredient in the assembly process is the instruction manual, which any robotic platform intended to do assembly tasks must be equipped to follow for safety and physical feasibility. In \datasetName, for each assembly step and each camera view, we render the diagram using Blender as follows. First, we use a line-art style without coloring to mimic the visual style in real-world instruction manuals such as IKEA's. Then, we use the CAD part position at the final time step of the trajectory to represent the part's final assembled state. We also render a segmentation map of this diagram for later text annotations.

We render the diagrams using a fixed set of isometric camera views. The diagrams from multiple camera views are used in two ways. First, they are selected in the later text annotation stage as the reference materials for a large vision-and-language model (VLM), 
as detailed later. Second, they are used to choose the camera view for the instruction manual, which uses a fixed camera view throughout all assembly steps  
(see Supplementary Material for details).
 
\subsubsection{Instructional Text Generation:}
\label{sec:text-gen}

To construct \datasetName, we generate text instructions to accompany each diagram, forming realistic step-by-step manuals. Our pipeline has two stages. First, we prompt a VLM (GPT-4.1) with diagrams of all individual parts to assign consistent names (e.g., ``fastener'', ``wire frame''), ensuring uniform terminology throughout the manual. Second, using these names, we prompt the VLM to produce textual instructions for each assembly step based on that step’s diagram.

Because our industrial assemblies contain complex shapes, frequent occlusions, and repeated part types (e.g., multiple identical screws), obtaining consistent part names from a single camera view is challenging. Parts may be hidden in later steps (temporal occlusion), blocked from view (spatial occlusion), or duplicated. To address these issues, we apply several visual prompting techniques (described in the Supplementary Material). After generating consistent part names (see Fig.~\ref{fig:part-name}), we produce the step-by-step instructions by providing the VLM with: (i) the current step’s diagram with the target part highlighted, and (ii) the same diagram with all parts color-coded and labeled. This ensures consistent naming across all assembly steps (see “Add Text Instructions” in Fig.~\ref{fig:dataset}).

\subsection{\modelName: Our Assembly Model}
\label{sec:model}

\begin{figure*}
    \centering
    \includegraphics[width=.8\linewidth]{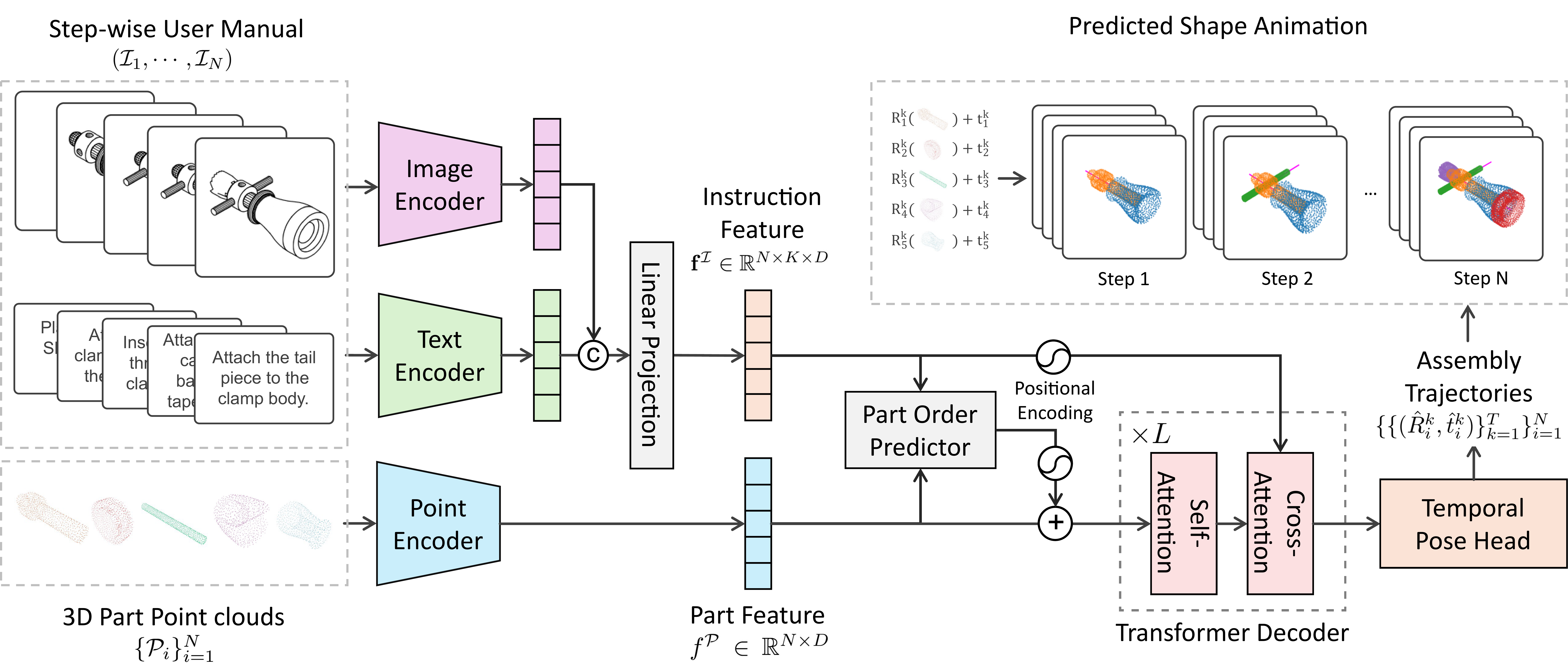}
    \vspace{-6pt}
    \caption{\textbf{Model Architecture of \modelName}. (1) Feature Extraction: \modelName starts with multi-modal encoders, converting user manual instructions and 3D part point clouds into embeddings of the same feature dimension $D$. (2) Predict Part Order: we use an existing predictor to get part order in the form of a permutation matrix. (3) Predict Assembly Trajectories: we use a transformer decoder with positional encodings to predict the assembly trajectory for each step. }
\label{fig:architecture}
\vspace{-3mm}
\end{figure*}

As shown in Figure~\ref{fig:architecture}, the input to \modelName is a step-by-step instruction manual and the corresponding set of separated 3D part point clouds. Our model starts with multimodal encoders, converting each instruction step and each part's point cloud into feature embeddings of the same dimension $D$. After using an existing predictor to obtain part orders in the form of a permutation matrix, we apply a transformer decoder with positional encodings to predict the assembly trajectory for each step, as explained below.

\subsubsection{Feature Extraction:}
We begin by applying off-the-shelf encoders to obtain semantic latent features from the inputs. For the 3D part point clouds $\{\mathcal{P}_i\}_{i=1}^N$, we use a lightweight PointNet variant~\cite{qi2017pointnet}, similar to that in~\cite{li2020imagepa,zhang2025manual}.
For the sequence of step diagrams $(I_j^{\mathrm{img}})_{j=1}^N$, since the assembly instructions progress incrementally, we focus on the differences between successive steps. For any pair of consecutive diagrams $\mathcal{I}_j^{\mathrm{img}}$ and $\mathcal{I}_{j+1}^{\mathrm{img}}$, $j \in \{1, 2, \dots, N-1$\}, we form a difference image $\lvert \mathcal{I}_j^{\mathrm{img}} - \mathcal{I}_{j+1}^{\mathrm{img}} \rvert$ that highlights the newly added part relative to the partially assembled object. This difference image is then divided into $K$ patches and passed through a DINOv3 image encoder~\cite{siméoni2025dinov3} to extract features. 
For the corresponding text content, we use the Qwen-3 embedding model~\cite{zhang2025qwen3embedding} with frozen model weights. Each of the three modalities is projected to the same feature dimensionality $D$ using linear layers, yielding part features $f^{\mathcal{P}} \in \mathbb{R}^{N \times D}$, image features $f^{\mathrm{img}} \in \mathbb{R}^{N \times K \times D},$ and text features $f^{\mathrm{txt}} \in \mathbb{R}^{N \times D}.$ To fuse image and text features into one feature, we concatenate the two features by repeating text features along the patch dimension and apply linear projections,
yielding instruction features $\mathbf{f}^\mathcal{I} \in \mathbb{R}^{N \times K \times D}$.

\subsubsection{Predicting Part Assembly Order:}
Applying a similar approach as Manual-PA~\cite{zhang2025manual}, we calculate a similarity matrix between the embeddings of the parts and those of the instructions with a max-pooling operation on patch dimension $K$. Then, we use the Hungarian matching algorithm~\cite{kuhn1955Hungarian} to convert the similarity matrix into the predicted order $(\hat{\order}_1, \hat{\order}_2, \cdots, \hat{\order}_N)$, which is projected into a permutation matrix $\permuteMat \in \{0,1\}^{N \times N}$.

\subsubsection{Predicting Assembly Trajectories:}
To perform part-to-part interactions, we add the permutation matrix $\permuteMat$ with positional encoding~\cite{vaswani2023attentionneed} to the part features $f^P$ and send the results to a self-attention transformer decoder. Then, we 
feed the outputs to a cross-attention module with temporal dimension, where position-encoded instruction features are added to produce the latent feature of assembly trajectories, shaped as $\mathbb{R}^{N \times T \times D}$. Finally, the latent feature is converted into a sequence of poses $\bigl(\bigl\{\{(\hat{R}_i^k, \hat{t}_i^k) \}_{k=1}^T\bigr\}_{i=1}^N\bigr)$ using a pose prediction head~\cite{zhang2025manual}, where the rotations are represented using quaternions.

\subsubsection{Training Losses:}
We train one model with the above architecture for part order prediction, and a second model with the same architecture for trajectory prediction. During trajectory prediction learning, we always feed the model using the ground-truth part order. 

\subsubsection{Loss for Order Prediction:}
We optimize the similarity matrix between the instruction features $f_j^\mathcal{I}$ and part features $f_{\sigma(i)}^\mathcal{P}$. 
In the similarity matrix, a correct match between $f_j^\mathcal{I}$ and $f_{\sigma(i)}^\mathcal{P}$ yields a higher value.
Based on this motivation, we adopt an InfoNCE loss \cite{hadsell206,oord2019representation} design:
\begin{equation}
\mathcal{L}_{\text{order}}
= - \frac{1}{B} \sum_{i=1}^{B}
\log \frac{
    \exp\!\left( \mathrm{sim}( \mathbf{f}^{\mathcal{P}}_{\sigma(i)}, \mathbf{f}^{\mathcal{I}}_{i} ) / \tau \right)
}{
    \sum_{j=1}^{B}
    \exp\!\left( \mathrm{sim}( \mathbf{f}^{\mathcal{P}}_{\sigma(i)}, \mathbf{f}^{\mathcal{I}}_{j} ) / \tau \right)\notag
},
\label{eq:order}
\end{equation}
where $B$ is the batch size, $\sigma(i)$ denotes a permutation over part indices, $\mathrm{sim}(\cdot)$ computes the similarity between two features, and $\tau$ is a temperature scaling factor.

\subsubsection{Loss for Trajectory Prediction:}

Inspired by~\cite{zhang2025manual,li2020imagepa}, the point-cloud loss $\mathcal{L}_P$ measures the difference between the point clouds of the predicted final assembly at step $T$ (all parts in their final poses) and the ground-truth assembly.
\vspace{-2mm}
\begin{equation}
\mathcal{L}_P = 
\mathrm{CD}\!\left(
\bigcup_{i=1}^{N} (\hat{R}_{i}^{(T)} P_i + \hat{t}_{i}^{(T)}),
\;
\bigcup_{i=1}^{N} (R_i^{(T)} P_i + t_i^{(T)})
\right),\notag
\label{eq:9_time}
\end{equation}
where $\bigcup_{i=1}^{N}$ represents the union of all $N$ transformed parts to form the complete assembled shape, and $\mathrm{CD}(\cdot)$ denotes the bidirectional chamfer distance, which measures the difference between two point clouds~\cite{fan2016pointsetgenerationnetwork}.

In addition to the overall point-cloud loss, we separately measure  translation and rotation losses. For translation loss,
\vspace{-2mm}
\begin{equation}
\mathcal{L}_T = \frac{1}{NT} \sum_{i=1}^{N} \sum_{k=1}^{T} \lVert \hat{t}_{i}^{(k)} - t_i^{(k)} \rVert_2,\notag
\label{eq:6_time}
\end{equation}
\noindent where $\hat{t}_{i}^{(k)}$ is the predicted translation at time step $k$, $t_i$ is the ground-truth translation, and $\lVert \cdot \rVert_2$ denotes the $\ell_2$ norm.

For rotational loss, as parts may have rotational symmetries, solely relying on $\ell_2$ distance will miss some correct answers. So we use chamfer distance on the point clouds,
\vspace{-2mm}
\begin{equation}
\mathcal{L}_R = \frac{1}{NT} \sum_{i=1}^{N} \sum_{k=1}^{T} \mathrm{CD}(\hat{R}_{i}^{(k)} P_i, R_i^{(k)} P_i),\notag
\label{eq:7_time}
\end{equation}
where $\hat{R}_{i}^{(k)}$ is the predicted rotation for part $i$ at time step $k$, $R_i$ is the ground-truth rotation, and $\mathrm{CD}(\cdot)$ denotes the bidirectional chamfer distance between the rotated point clouds.

To encourage temporally smooth motions, we additionally regularize the frame-to-frame ``velocity'' of both translations and rotations of the predicted trajectories. We penalize the finite difference between consecutive frames,
\begin{align}
\mathcal{L}_{S_T}
= \frac{1}{N(T-1)} \sum_{i=1}^{N} \sum_{k=1}^{T-1}
\big\lVert \hat{t}_{i}^{(k+1)} - \hat{t}_{i}^{(k)} \big\rVert_2^2, 
\label{eq:smooth_loss_trans} \\
\mathcal{L}_{S_R}
= \frac{1}{N(T-1)} \sum_{i=1}^{N} \sum_{k=1}^{T-1}
\big\lVert \hat{q}_{i}^{(k+1)} - \hat{q}_{i}^{(k)} \big\rVert_2^2.
\label{eq:smooth_loss_quat}
\end{align}

The final loss is a weighted sum of all the components:
\begin{equation}
\mathcal{L}
= \lambda_P \mathcal{L}_P
+ \lambda_T \mathcal{L}_T
+ \lambda_R \mathcal{L}_R
+ \lambda_{S_T} \mathcal{L}_{S_T}
+ \lambda_{S_R} \mathcal{L}_{S_R}.
\label{eq:10_time}
\end{equation}

\subsection{Physics-Aware Evaluation in \datasetName}

\label{sec:eval}

Our evaluation captures three aspects of assembly: if a model: i) correctly grounds each diagram to its part, ii) can make correct prediction of the final 3D poses of the parts (\emph{Static Pose Estimate}), and iii) can predict 3D assembly trajectories that are physically feasible (\emph{Assembly in Simulator}). For (i), we use the standard Kendall's Tau (KD)~\cite{kendall} metric to compute the correlation between the actual assembly order $(\order_1, \order_2, \cdots,\order_N)$ and the predicted one $(\hat{\order}_1, \hat{\order}_2, \cdots,\hat{\order}_N)$. For (ii), similar to prior works, ~\cite{zhang2025manual,li2020imagepa,pmlr-v229-li23a}, we use the Shape Chamfer Distance (SCD), Part Assembly Correctness (PA), and Success Rate (SR). For (iii), we detail our new evaluation framework below.

As noted above, physical feasibility of predicted assembly trajectories is important for real-world adoption. In order to capture such feasibility,  we propose to execute the predicted trajectories within a physics-based simulator to check their correctness. We use Newton Physics~\cite{newton2025} as our simulator and evaluate the assembly process step-by-step: For each predicted step, we use the velocity sequence of the predicted trajectory as the control signal to roll out the simulation. 
Then we measure the difference between the simulation results and the ground truth.

\begin{figure}
    \centering
    \includegraphics[width=.9\linewidth]{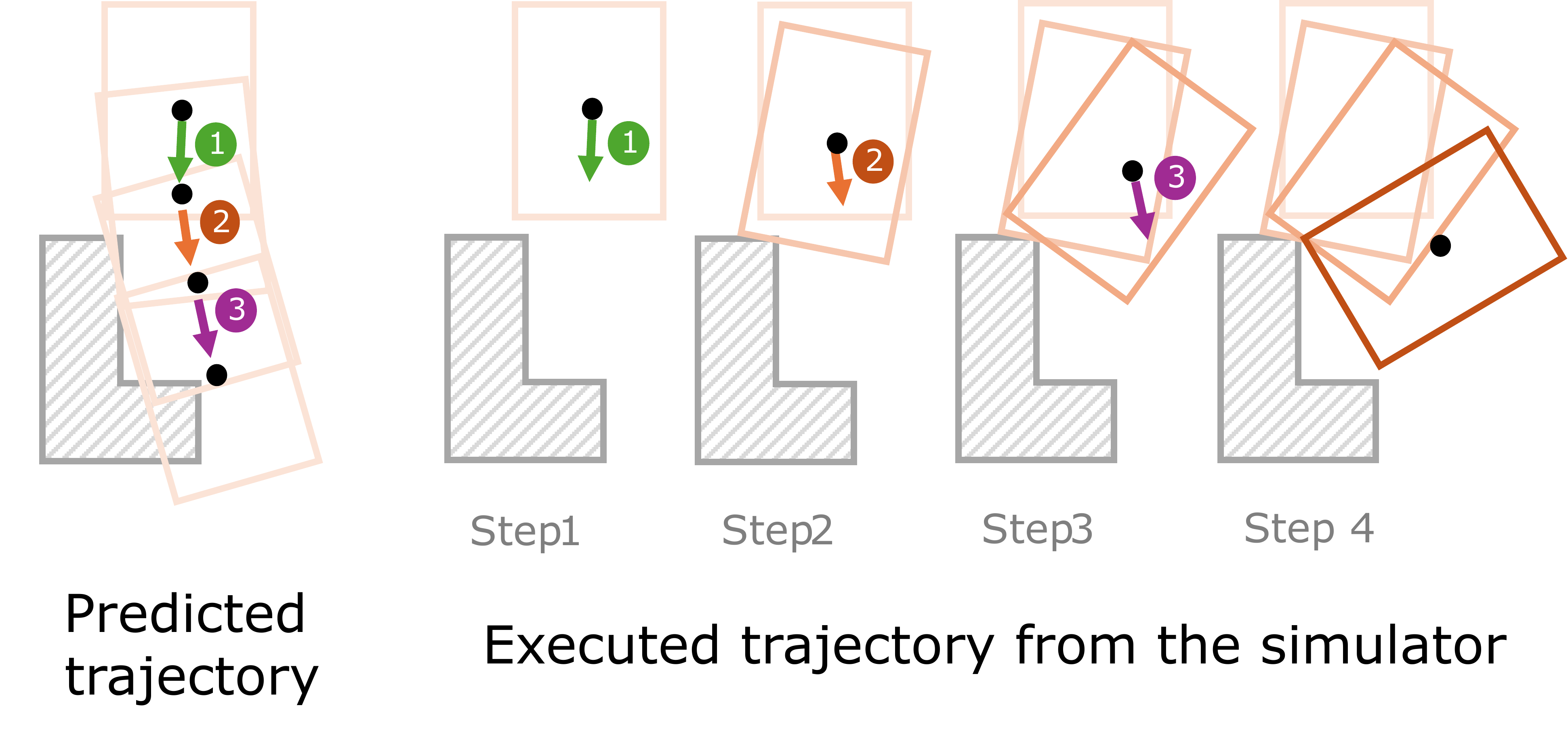}
    \vspace{-6pt}
    \caption{We use a physics simulator to execute the predicted assembly trajectory, evaluating whether it is physically feasible.}
    \label{fig:sim-explain}
    \vspace{-3mm}
\end{figure}

Initially, we set up the parts that were assembled in previous steps of the instruction manual within the simulator, using their predicted final poses. Then we place the current part that is to be assembled, using the pose from the first time step of its predicted trajectory. We ignore gravity in this work for simplicity. 
Next, as shown in Figure~\ref{fig:sim-explain}, we let $\Delta t$ represent the duration of each time step of the trajectory.
Then we start the simulation by assigning $v_1$ as the initial velocity of the part and let the simulation run for $\Delta t$. The part might collide into other parts and change its velocity during that time step. After $\Delta t$, we iteratively assign $v_2$ and $v_3$ each for the same amount of time, and so on through to the final time step. The process is depicted in Figure~\ref{fig:sim-explain}.

We measure the prediction quality by comparing the executed pose trajectory from the simulator to the ground truth. Apart from using \textbf{PA} and \textbf{SR} in the simulation setting by following the same algorithms, we also present two new metrics to account for part symmetries. Specifically, due to the part symmetries and thus potential for non-unique solution poses, we do not compute the translation or rotational differences. Instead, we apply chamfer distance on two common trajectory evaluation metrics, i.e., ADE and FDE, resulting in the following variants:
\begin{itemize}
    \item \textbf{Average Chamfer Distance (ACD)}: For each time step, we apply the corresponding executed pose and ground truth pose to the point clouds of their related parts and computes the chamfer distance between them. The chamfer distances from all time steps are averaged and the quartile values for all parts in all assemblies is reported.
    \item \textbf{Final Chamfer Distance (FCD)}: Similar to ACD, but instead of averaging over all time steps, we only take the chamfer distance of the final time step and report the median over all assemblies. We report the quartile values for all parts in all shapes.
\end{itemize}

\begin{table*}[!ht]
    \footnotesize
    \caption{\textbf{Part assembly results on the test split of \datasetName}. We bold the best results and highlight the second best results in blue.}
    \vspace{-6pt}
    \centering
    \setlength{\tabcolsep}{3pt}
    \begin{tabular*}{\textwidth}{@{\extracolsep{\fill}}lcccccccccccc@{}}
        \toprule
        \multirow{2}{*}{\textbf{Model}} 
            & \textbf{Ordering} 
            & \multicolumn{3}{c}{\textbf{Final Pose Estimate}}
            & \multicolumn{8}{c}{\textbf{Assembly in Physics Simulator}} \\
        \cmidrule(lr){2-2}
        \cmidrule(lr){3-5}
        \cmidrule(lr){6-13}
            & KD$\uparrow$
            & SCD($10^{-3}$)$\downarrow$
            & PA(\%)$\uparrow$
            & SR(\%)$\uparrow$
            & \multicolumn{3}{c}{ACD($10^{-3}$)$\downarrow$}
            & \multicolumn{3}{c}{FCD($10^{-3}$)$\downarrow$} 
            & PA(\%)$\uparrow$
            & SR(\%)$\uparrow$\\
        \cmidrule(lr){6-8}
        \cmidrule(lr){9-11}
            & 
            & 
            & 
            & 
            & 25\% & 50\% & 75\%
            & 25\% & 50\% & 75\% 
            & 
            & \\
        \midrule

        \multicolumn{2}{l}{\textit{Standard Setting}} \\
        \textbf{\modelName}
            & \textbf{0.819}
            & \textcolor{blue}{3.91}
            & \textbf{71.21}
            & \textcolor{blue}{34.64}
            & \textbf{6.95} & \textcolor{blue}{34.77} & \textcolor{blue}{156.74}
            & \textcolor{blue}{3.31} & \textbf{15.97} & \textcolor{blue}{110.30}
            & \textbf{42.76}
            & \textcolor{blue}{13.57}\\
        w/o text
            & \textcolor{blue}{0.805}
            & \textbf{3.83}
            & \textcolor{blue}{70.28}
            & \textbf{35.00}
            & \textcolor{blue}{6.99} & \textbf{33.68} & \textbf{148.86}
            & \textbf{3.03} & \textcolor{blue}{17.25} & \textbf{99.77}
            & \textcolor{blue}{41.77}
            & \textbf{15.00}\\
        w/o trajectory
            & \textbf{0.819}
            & 4.42
            & 67.63
            & 30.00
            & 104.73 & 188.94 & 317.76
            & 11.95 & 58.52 & 199.64
            & 22.09
            & 1.43\\
        ManualPA~\cite{zhang2025manual} (ICCV'25)
            & 0.788
            & 4.24
            & 70.04
            & 33.57
            & 104.66 & 192.79 & 325.50
            & 11.75 & 56.04 & 209.09 
            & 23.24
            & 1.79\\

        \midrule

        \multicolumn{2}{l}{\textit{Use GT part orders}} \\
        \textbf{\modelName}
            & --
            & \textcolor{blue}{3.87}
            & \textbf{79.69}
            & \textbf{44.29}
            & \textbf{3.23} & \textbf{9.97} & \textbf{29.74}
            & \textbf{1.50} & \textbf{4.43} & \textbf{12.41}
            & \textbf{70.15}
            & \textbf{33.57}\\
        w/o text
            & --
            & \textbf{3.78}
            & 78.19
            & 42.86
            & \textcolor{blue}{3.45} & \textcolor{blue}{10.77} & \textcolor{blue}{35.28}
            & \textcolor{blue}{1.61} & \textcolor{blue}{4.85} & \textcolor{blue}{14.32}
            & \textcolor{blue}{67.96}
            & \textcolor{blue}{31.79}\\
        w/o trajectory
            & --
            & 3.80
            & \textcolor{blue}{79.31}
            & \textcolor{blue}{43.21}
            & 136.48 & 235.16 & 390.23
            & 6.90 & 37.39 & 197.65
            & 29.85
            & 3.57\\
        ManualPA~\cite{zhang2025manual} (ICCV'25)
            & --
            & 4.15
            & 77.40
            & 39.28
            & 142.32 & 230.52 & 382.62
            & 6.02 & 35.00 & 196.51
            & 31.33
            & 2.14\\
        \bottomrule
    \end{tabular*}

    \label{tab:main-metrics}
    \vspace{-3mm}
\end{table*}


\section{Experiments}
\label{sec:experiment}

Table~\ref{tab:main-metrics} provides a comprehensive evaluation of assembly trajectory prediction, final pose estimation, and full assembly simulation across multiple ablations. We compare \modelName to its ablations and to \cite{zhang2025manual}. Since some baselines (e.g.,~\cite{zhang2025manual}) do not predict trajectories, we generate them heuristically: from the predicted final pose, we translate the part outward from the object's center of mass (without rotation) for a distance equal to half the diagonal of its predicted bounding-box. Reversing this path yields the assembly motion. 

\begin{figure}
    \centering
    \includegraphics[width=.9\linewidth]{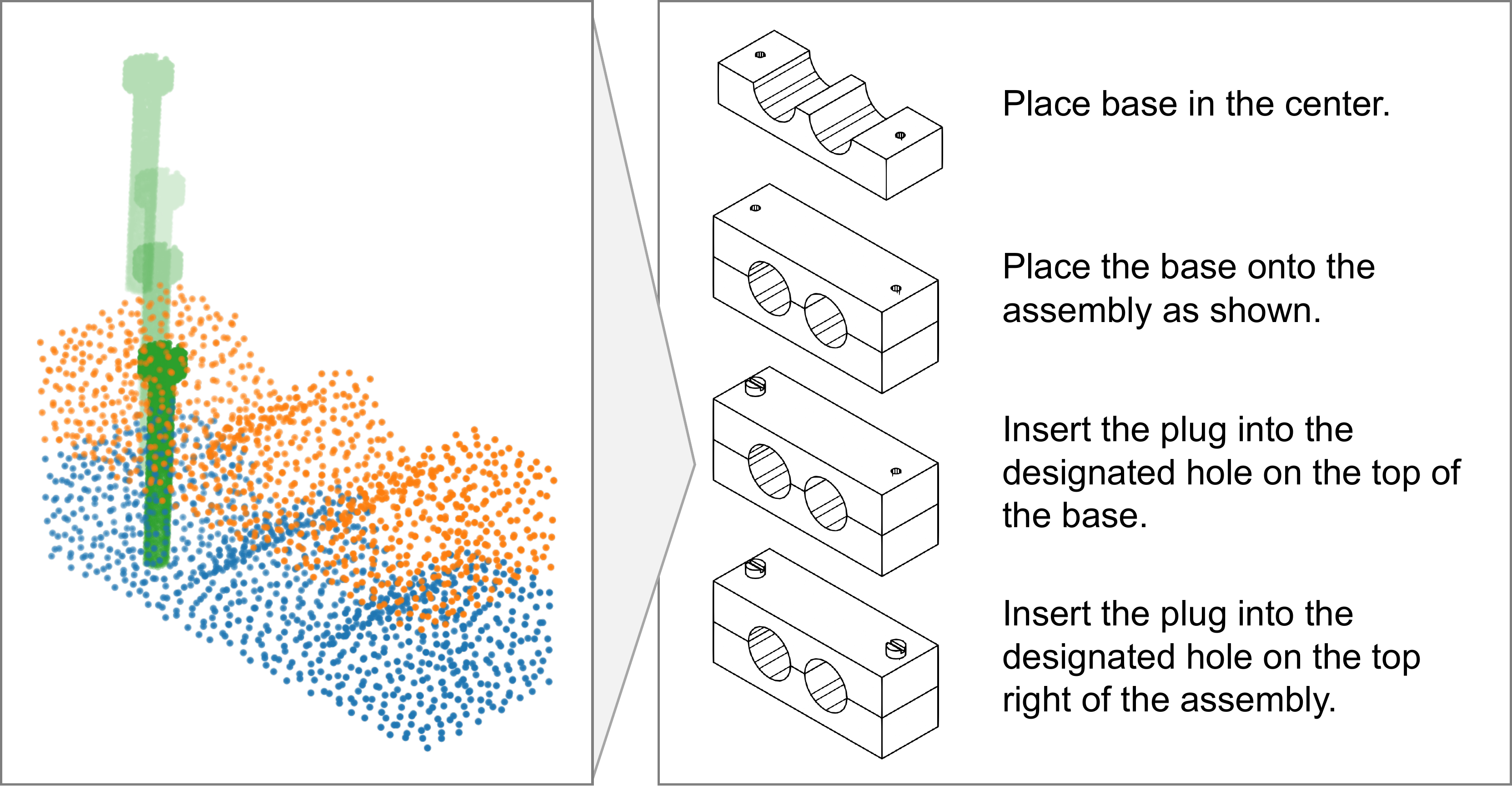}
    \vspace{-6pt}
    \caption{Given the instruction manual on the right, \modelName showcases the capability of predicting insertion assembly trajectory. Key frames of the trajectory are shown on the left.}
    \label{fig:assm-pred}
    \vspace{-4mm}
\end{figure}

\noindent\textbf{Part Order and Final Pose:} We first analyze the \emph{GT part-order} setting, which serves as a sanity check to isolate the model’s trajectory-prediction and pose-estimation capabilities from ordering errors. When provided with perfect part orders, \modelName consistently achieves the strongest performance across nearly all me  trics. In the \emph{Final Pose Estimate} block, AssemblyDyno attains the lowest SCD scores and the highest PA/SR values among the compared methods, highlighting its accurate trajectory reasoning and robust geometric understanding. In the \emph{Assembly in Simulator} block, AssemblyDyno also yields superior PA and SR, even though the model is not trained with simulator feedback (see Figure~\ref{fig:assm-pred}). This demonstrates strong real-to-sim transfer of predicted motion trajectories. The ablation results further validate key model components: removing the text encoder (\texttt{w/o text}) or replacing the trajectory predictor with the naïve baseline (\texttt{w/o trajectory}) leads to consistent drops in PA, SR, ACD, and FCD, underscoring the importance of both textual grounding and the dedicated trajectory module. 

 We then analyze the \emph{standard setting}, where the model must predict part orders before generating trajectories. Here, performance decreases across all methods, reflecting the substantial influence of ordering quality on downstream assembly outcomes. Despite this, \modelName still outperforms prior work, though the margin over its own ablations becomes smaller. This trend is especially visible in PA and SR, where the advantage of \modelName over \texttt{w/o text} narrows. We attribute this to two key factors: (1) assembly order prediction becomes the dominant bottleneck, and errors in ordering propagate into trajectory and pose predictions; and (2) textual information contributes limited additional signal for order prediction, since step-wise diagrams already strongly constrain the next operation, reducing the marginal benefit of language inputs at this stage. Overall, the table highlights three central insights: (i) our model is intrinsically strong at pose and trajectory prediction, as shown by the GT-order results; (ii) the ablations validate the contributions of the text encoder and structured trajectory module; 
and (iii) in realistic deployment conditions, improving part-order prediction is crucial for unlocking further gains in downstream assembly performance.

\noindent\textbf{Physics-based Evaluation:}

\begin{figure}
    \centering
    \includegraphics[width=.8\linewidth]{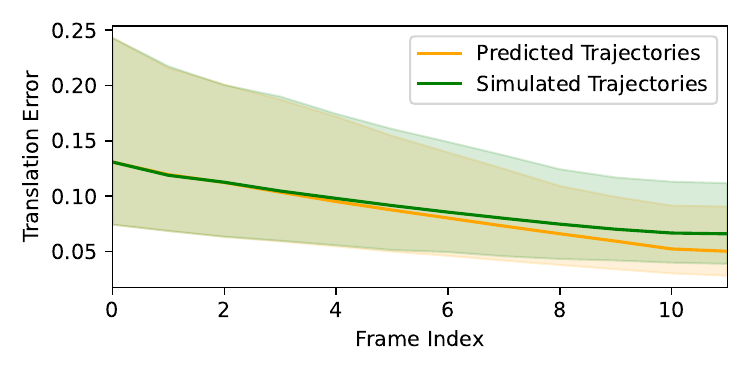}
    \vspace{-3mm}
    \caption{Median translation error with respect to ground truth as a function of trajectory frame. Shaded regions denote the 25th–75th percentiles computed across all object parts. 
    The simulated trajectories exhibit consistently larger deviations due to collisions and other dynamic interactions, revealing hidden failure modes that are not captured when evaluating predictions alone. 
    }
    \label{fig:deviance_sim}
    \vspace{-5mm}
\end{figure}

We further show that our simulator-based evaluation provides a more stringent assessment of trajectory quality. Figure~\ref{fig:deviance_sim} compares the median translation errors of the \emph{predicted} trajectories (orange) and the corresponding \emph{simulated} trajectories (green). Although both decrease over time, the simulated error remains higher and the gap widens toward later frames, indicating that the simulator exposes compounding inaccuracies such as collisions or infeasible motions. Early in the trajectory, the simulated-error variance is larger because early-time-step predictions contains large variance that deviate substantially from ground truth. As the object approaches assembly, the physical constraints align more closely with the true configuration, reducing variance; however, these same constraints and obstacles still impede the predicted motion from reaching the final pose, resulting in a slightly higher but more stable residual error. Thus, the simulator-based metric offers a stricter and more realistic measure of trajectory feasibility.

\section{Conclusion}
\label{sec:conclusion}

We introduced \datasetName, a large-scale, multi-modal assembly dataset that extends beyond furniture to include complex industrial objects, complete with step-wise diagrams, textual descriptions, and ground-truth 6-DoF part trajectories. Building on this foundation, we presented \modelName, a unified transformer-based architecture that jointly predicts assembly order, final poses, and physically plausible motion trajectories. Our experiments demonstrate that existing state-of-the-art methods struggle on the richer and more challenging scenarios offered by \datasetName, while \modelName achieves substantially stronger performance in both final pose estimation and simulator-executed assembly. The analysis further highlights the benefits of integrating multi-modal manual information and structured trajectory prediction, as well as the importance of accurate order prediction for full assembly success. Overall, our work provides a comprehensive benchmark and modeling framework that brings instruction-guided assembly significantly closer to real-world applicability, with opportunities for future advances in order prediction, physical reasoning, and robotic execution.

Please see the supplementary material for more detailed results.


{
    \small
    \bibliographystyle{ieeenat_fullname}
    \bibliography{main}

@String(CVPR= {IEEE Conf. Comput. Vis. Pattern Recog.})

@String(ICCV= {Int. Conf. Comput. Vis.})

@String(ECCV= {Eur. Conf. Comput. Vis.})

@String(NIPS= {Adv. Neural Inform. Process. Syst.})

@String(CVPR  = {CVPR})

@String(ICCV  = {ICCV})

@String(ECCV  = {ECCV})

@String(NIPS  = {NeurIPS})

@ARTICLE{kavraki1996,
  author={Kavraki, L.E. and Svestka, P. and Latombe, J.-C. and Overmars, M.H.},
  journal={IEEE Transactions on Robotics and Automation}, 
  title={Probabilistic roadmaps for path planning in high-dimensional configuration spaces}, 
  year={1996},
  volume={12},
  number={4},
  pages={566-580},
  doi={10.1109/70.508439}}

@article{LaValle1998RRT,
  title={Rapidly-exploring random trees : a new tool for path planning},
  author={Steven M. LaValle},
  journal={The annual research report},
  year={1998},
}

@article{tian2022assemble,
    title={Assemble Them All: Physics-Based Planning for Generalizable Assembly by Disassembly},
    author={Tian, Yunsheng and Xu, Jie and Li, Yichen and Luo, Jieliang and Sueda, Shinjiro and Li, Hui and Willis, Karl D.D. and Matusik, Wojciech},
    journal={ACM Trans. Graph.},
    volume={41},
    number={6},
    articleno={278},
    numpages={15},
    year={2022},
    publisher={ACM}
}

@inproceedings{tian2024asap,
  title={Asap: Automated sequence planning for complex robotic assembly with physical feasibility},
  author={Tian, Yunsheng and Willis, Karl DD and Al Omari, Bassel and Luo, Jieliang and Ma, Pingchuan and Li, Yichen and Javid, Farhad and Gu, Edward and Jacob, Joshua and Sueda, Shinjiro and others},
  booktitle={2024 IEEE International Conference on Robotics and Automation (ICRA)},
  pages={4380--4386},
  year={2024},
  organization={IEEE}
}

@article{kingston2018review,
   author = "Kingston, Zachary and Moll, Mark and Kavraki, Lydia E.",
   title = "Sampling-Based Methods for Motion Planning with Constraints", 
   journal= "Annual Review of Control, Robotics, and Autonomous Systems",
   year = "2018",
   volume = "1",
   number = "Volume 1, 2018",
   pages = "159-185",
   doi = "https://doi.org/10.1146/annurev-control-060117-105226",
   url = "https://www.annualreviews.org/content/journals/10.1146/annurev-control-060117-105226",
   publisher = "Annual Reviews",
   issn = "2573-5144",
   type = "Journal Article",
  }

@inproceedings{li2020imagepa,
author = {Li, Yichen and Mo, Kaichun and Shao, Lin and Sung, Minhyuk and Guibas, Leonidas},
title = {Learning 3D Part Assembly from a Single Image},
year = {2020},
isbn = {978-3-030-58538-9},
publisher = {Springer-Verlag},
address = {Berlin, Heidelberg},
url = {https://doi.org/10.1007/978-3-030-58539-6_40},
doi = {10.1007/978-3-030-58539-6_40},
booktitle = {Computer Vision – ECCV 2020: 16th European Conference, Glasgow, UK, August 23–28, 2020, Proceedings, Part VI},
pages = {664–682},
numpages = {19},
location = {Glasgow, United Kingdom}
}

@inproceedings{zhang2025manual,
  title={{Manual-PA}: Learning 3d part assembly from instruction diagrams},
  author={Zhang, Jiahao and Cherian, Anoop and Rodriguez, Cristian and Deng, Weijian and Gould, Stephen},
  booktitle={Proceedings of the IEEE/CVF International Conference on Computer Vision},
  pages={6304--6314},
  year={2025}
}

@inproceedings{checkmanual,
  author    = {Long, Yuxing and Zhang, Jiyao and Pan, Mingjie and Wu, Tianshu and Kim, Taewhan and Dong, Hao},
  title     = {{CheckManual}: A New Challenge and Benchmark for Manual-based Appliance Manipulation},
  booktitle = {Proceedings of the IEEE/CVF Conference on Computer Vision and Pattern Recognition (CVPR)},
  month     = {June},
  year      = {2025},
}

@inproceedings{text2cad,
	author = {Khan, Mohammad Sadil and Sinha, Sankalp and Sheikh, Talha Uddin and Stricker, Didier and Ali, Sk Aziz and Afzal, Muhammad Zeshan},
	booktitle = {Advances in Neural Information Processing Systems},
	editor = {A. Globerson and L. Mackey and D. Belgrave and A. Fan and U. Paquet and J. Tomczak and C. Zhang},
	pages = {7552--7579},
	publisher = {Curran Associates, Inc.},
	title = {Text2CAD: Generating Sequential CAD Designs from Beginner-to-Expert Level Text Prompts},
	url = {https://proceedings.neurips.cc/paper_files/paper/2024/file/0e5b96f97c1813bb75f6c28532c2ecc7-Paper-Conference.pdf},
	volume = {37},
	year = {2024}}

@article{xu2025cadmllm,
  title={{CAD-MLLM}: Unifying multimodality-conditioned {CAD} generation with {MLLM}},
  author={Xu, Jingwei and Wang, Chenyu and Zhao, Zibo and Liu, Wen and Ma, Yi and Gao, Shenghua},
  journal={arXiv preprint arXiv:2411.04954},
  year={2024}
}

@article{kendall,
  title={A new measure of rank correlation},
  author={Kendall, Maurice G},
  journal={Biometrika},
  volume={30},
  number={1-2},
  pages={81--93},
  year={1938},
  publisher={Oxford University Press}
}

@InProceedings{pmlr-v229-li23a,
  title = 	 {Rearrangement Planning for General Part Assembly},
  author =       {Li, Yulong and Zeng, Andy and Song, Shuran},
  booktitle = 	 {Proceedings of The 7th Conference on Robot Learning},
  pages = 	 {127--143},
  year = 	 {2023},
  editor = 	 {Tan, Jie and Toussaint, Marc and Darvish, Kourosh},
  volume = 	 {229},
  series = 	 {Proceedings of Machine Learning Research},
  month = 	 {06--09 Nov},
  publisher =    {PMLR},
  url = 	 {https://proceedings.mlr.press/v229/li23a.html},
}

@InProceedings{cadllama2025,
    author    = {Li, Jiahao and Ma, Weijian and Li, Xueyang and Lou, Yunzhong and Zhou, Guichun and Zhou, Xiangdong},
    title     = {{CAD-Llama}: Leveraging Large Language Models for Computer-Aided Design Parametric 3D Model Generation},
    booktitle = {Proceedings of the IEEE/CVF Conference on Computer Vision and Pattern Recognition (CVPR)},
    month     = {June},
    year      = {2025},
    pages     = {18563-18573}
}

@InProceedings{Sener_2022_Assembly101,
    author    = {Sener, Fadime and Chatterjee, Dibyadip and Shelepov, Daniel and He, Kun and Singhania, Dipika and Wang, Robert and Yao, Angela},
    title     = {Assembly101: A Large-Scale Multi-View Video Dataset for Understanding Procedural Activities},
    booktitle = {Proceedings of the IEEE/CVF Conference on Computer Vision and Pattern Recognition (CVPR)},
    month     = {June},
    year      = {2022},
    pages     = {21096-21106}
}

@INPROCEEDINGS{EPIC-Tent,
  author={Jang, Youngkyoon and Sullivan, Brian and Ludwig, Casimir and Gilchrist, Iain D. and Damen, Dima and Mayol-Cuevas, Walterio},
  booktitle={2019 IEEE/CVF International Conference on Computer Vision Workshop (ICCVW)}, 
  title={EPIC-Tent: An Egocentric Video Dataset for Camping Tent Assembly}, 
  year={2019},
  volume={},
  number={},
  pages={4461-4469},
  doi={10.1109/ICCVW.2019.00547}}

@inproceedings{Zhang2023Aligning,
  author    = {Zhang, Jiahao and Cherian, Anoop and Liu, Yanbin and Ben-Shabat, Yizhak and Rodriguez, Cristian and Gould, Stephen},
  title     = {Aligning Step-by-Step Instructional Diagrams to Video Demonstrations},
  booktitle = {Conference on Computer Vision and Pattern Recognition (CVPR)},
  year      = {2023},
}

@article{sliwowski2025reassemble,
  title={Reassemble: A multimodal dataset for contact-rich robotic assembly and disassembly},
  author={Sliwowski, Daniel and Jadav, Shail and Stanovcic, Sergej and Orbik, Jedrzej and Heidersberger, Johannes and Lee, Dongheui},
  journal={arXiv preprint arXiv:2502.05086},
  year={2025}
}

@article{tie2025manual2skill,
  title={{Manual2skill}: Learning to read manuals and acquire robotic skills for furniture assembly using vision-language models},
  author={Tie, Chenrui and Sun, Shengxiang and Zhu, Jinxuan and Liu, Yiwei and Guo, Jingxiang and Hu, Yue and Chen, Haonan and Chen, Junting and Wu, Ruihai and Shao, Lin},
  journal={arXiv preprint arXiv:2502.10090},
  year={2025}
}

@article{tian2025fabrica,
  title={{Fabrica:} Dual-Arm Assembly of General Multi-Part Objects via Integrated Planning and Learning},
  author={Tian, Yunsheng and Jacob, Joshua and Huang, Yijiang and Zhao, Jialiang and Gu, Edward and Ma, Pingchuan and Zhang, Annan and Javid, Farhad and Romero, Branden and Chitta, Sachin and others},
  journal={arXiv preprint arXiv:2506.05168},
  year={2025}
}

@inproceedings{wang2022ikea,
  title={{IKEA-Manual}: Seeing Shape Assembly Step by Step},
  author={Wang, Ruocheng and Zhang, Yunzhi and Mao, Jiayuan and Zhang, Ran and Cheng, Chin-Yi and Wu, Jiajun},
  booktitle={NeurIPS 2022 Datasets and Benchmarks Track},
    year={2022}
}

@inproceedings{ben2021ikea,
  title={The ikea asm dataset: Understanding people assembling furniture through actions, objects and pose},
  author={Ben-Shabat, Yizhak and Yu, Xin and Saleh, Fatemeh and Campbell, Dylan and Rodriguez-Opazo, Cristian and Li, Hongdong and Gould, Stephen},
  booktitle={Proceedings of the IEEE/CVF Winter Conference on Applications of Computer Vision},
  pages={847--859},
  year={2021}
}

@inproceedings{
      liu2024ikea,
      title={{IKEA} Manuals at Work: 4D Grounding of Assembly Instructions on Internet Videos},
      author={Yunong Liu and Cristobal Eyzaguirre and Manling Li and Shubh Khanna and Juan Carlos Niebles and Vineeth Ravi and Saumitra Mishra and Weiyu Liu and Jiajun Wu},
      booktitle={The Thirty-eight Conference on Neural Information Processing Systems Datasets and Benchmarks Track},
      year={2024}
      }

@misc{newton2025,
  title = {{Newton}: {GPU}-accelerated physics simulation for robotics, and simulation research.},
  author = {{Newton Contributors}},
  year = {2025},
  url = {https://github.com/newton-physics/newton},
  organization = {{Newton a Series of LF Projects, LLC}},
  license = {Apache-2.0}
}

@book{hess2007essential,
  title={The essential Blender: guide to 3D creation with the open source suite Blender},
  author={Hess, Roland},
  year={2007},
  publisher={No Starch Press}
}

@article{tie2025manual2skillconnect,
  title={{Manual2Skill++}: Connector-Aware General Robotic Assembly from Instruction Manuals via Vision-Language Models},
  author={Tie, Chenrui and Sun, Shengxiang and Lin, Yudi and Wang, Yanbo and Li, Zhongrui and Zhong, Zhouhan and Zhu, Jinxuan and Pang, Yiman and Chen, Haonan and Chen, Junting and others},
  journal={arXiv preprint arXiv:2510.16344},
  year={2025}
}

@inproceedings{zheng2023havid,
author = {Zheng, Hao and Lee, Regina and Lu, Yuqian},
title = {HA-ViD: a human assembly video dataset for comprehensive assembly knowledge understanding},
year = {2023},
publisher = {Curran Associates Inc.},
address = {Red Hook, NY, USA},
booktitle = {Proceedings of the 37th International Conference on Neural Information Processing Systems},
articleno = {2930},
numpages = {13},
location = {New Orleans, LA, USA},
series = {NIPS '23}
}

@inproceedings{zhu2023multilevelassembly,
  title={Multi-level reasoning for robotic assembly: From sequence inference to contact selection},
  author={Zhu, Xinghao and Jha, Devesh K and Romeres, Diego and Sun, Lingfeng and Tomizuka, Masayoshi and Cherian, Anoop},
  booktitle={2024 IEEE international conference on robotics and automation (ICRA)},
  pages={816--823},
  year={2024},
  organization={IEEE}
}

@inproceedings{tang2024automate,
  author    = {Tang, Bingjie and Akinola, Iretiayo and Xu, Jie and Wen, Bowen and Handa, Ankur and Van Wyk, Karl and Fox, Dieter and S. Sukhatme, Gaurav and Ramos, Fabio and Narang, Yashraj},
  title     = {AutoMate: Specialist and Generalist Assembly Policies over Diverse Geometries},
  booktitle = {Robotics: Science and Systems},
  year      = {2024},
}

@article{patel2025dynamo,
      title={DYNAMO: Dependency-Aware Deep Learning Framework for Articulated Assembly Motion Prediction}, 
      author={Mayank Patel and Rahul Jain and Asim Unmesh and Karthik Ramani},
      year={2025},
      journal={2509.12430},
      archivePrefix={arXiv},
      primaryClass={cs.CV},
      url={https://arxiv.org/abs/2509.12430}, 
}

@INPROCEEDINGS{li2024joint,
  author={Li, Yichen and Mo, Kaichun and Duan, Yueqi and Wang, He and Zhang, Jiequan and Shao, Lin and Matusik, Wojciech and Guibas, Leonidas},
  booktitle={2024 IEEE/CVF Conference on Computer Vision and Pattern Recognition (CVPR)}, 
  title={Category-Level Multi-Part Multi-Joint {3D} Shape Assembly}, 
  year={2024},
  volume={},
  number={},
  pages={3281-3291},
  doi={10.1109/CVPR52733.2024.00316}}

@InProceedings{lego2plan,
author="Wang, Ruocheng
and Zhang, Yunzhi
and Mao, Jiayuan
and Cheng, Chin-Yi
and Wu, Jiajun",
editor="Avidan, Shai
and Brostow, Gabriel
and Ciss{\'e}, Moustapha
and Farinella, Giovanni Maria
and Hassner, Tal",
title="Translating a Visual {LEGO} Manual to a Machine-Executable Plan",
booktitle="European Conference on Computer Vision (ECCV)",
year="2022",
publisher="Springer Nature Switzerland",
pages="677--694",
}

@INPROCEEDINGS{spaformer,
  author={Xu, Boshen and Zheng, Sipeng and Jin, Qin},
  booktitle={International Conference on 3D Vision (3DV)}, 
  title={{SPAFormer}: Sequential {3D} Part Assembly with Transformers}, 
  year={2025},
  pages={1317-1327},
  doi={10.1109/3DV66043.2025.00125}}

@InProceedings{Pun_2025_BrickGPT,
    author    = {Pun, Ava and Deng, Kangle and Liu, Ruixuan and Ramanan, Deva and Liu, Changliu and Zhu, Jun-Yan},
    title     = {Generating Physically Stable and Buildable Brick Structures from Text},
    booktitle = {Proceedings of the IEEE/CVF International Conference on Computer Vision (ICCV)},
    month     = {October},
    year      = {2025},
    pages     = {14798-14809}
}

@inproceedings{compoNet,
title = "CompoNet: Learning to generate the unseen by part synthesis and composition",
author = "Nadav Schor and Oren Katzir and Hao Zhang and Daniel Cohen-Or",
note = "Publisher Copyright: {\textcopyright} 2019 IEEE.; 17th IEEE/CVF International Conference on Computer Vision, ICCV 2019 ; Conference date: 27-10-2019 Through 02-11-2019",
year = "2019",
publisher = "IEEE",
pages = "8758--8767",
booktitle = "IEEE Proceedings of the International Conference on Computer Vision, (ICCV)",
}

@InProceedings{Wu_2020_pqnet,
author = {Wu, Rundi and Zhuang, Yixin and Xu, Kai and Zhang, Hao and Chen, Baoquan},
title = {{PQ-NET}: A Generative Part Seq2Seq Network for 3D Shapes},
booktitle = {IEEE/CVF Conference on Computer Vision and Pattern Recognition (CVPR)},
year = {2020}
}

@article{huang2025legocobuild,
  title={Lego co-builder: exploring fine-grained vision-language modeling for multimodal Lego assembly assistants},
  author={Huang, Haochen and Pei, Jiahuan and Aliannejadi, Mohammad and Sun, Xin and Ahsan, Moonisa and Yu, Chuang and Ren, Zhaochun and Cesar, Pablo and Wang, Junxiao},
  journal={arXiv preprint arXiv:2507.05515},
  year={2025}
}

@article{hasegawa-etal-2025-promqa-assembly,
  title={ProMQA-Assembly: Multimodal Procedural QA Dataset on Assembly},
  author={Hasegawa, Kimihiro and Imrattanatrai, Wiradee and Asada, Masaki and Holm, Susan and Wang, Yuran and Zhou, Vincent and Fukuda, Ken and Mitamura, Teruko},
  journal={arXiv preprint arXiv:2509.02949},
  year={2025}
}

@InProceedings{Mo_2019_Partnet,
    author = {Mo, Kaichun and Zhu, Shilin and Chang, Angel X. and Yi, Li and Tripathi, Subarna and Guibas, Leonidas J. and Su, Hao},
    title = {{PartNet}: A Large-Scale Benchmark for Fine-Grained and Hierarchical Part-Level {3D} Object Understanding},
    booktitle = {The IEEE Conference on Computer Vision and Pattern Recognition (CVPR)},
    month = {June},
    year = {2019}
}

@INPROCEEDINGS{qi2017pointnet,
  author={Charles, R. Qi and Su, Hao and Kaichun, Mo and Guibas, Leonidas J.},
  booktitle={2017 IEEE Conference on Computer Vision and Pattern Recognition (CVPR)}, 
  title={{PointNet}: Deep Learning on Point Sets for {3D} Classification and Segmentation}, 
  year={2017},
  volume={},
  number={},
  pages={77-85},
  doi={10.1109/CVPR.2017.16}}

@article{siméoni2025dinov3,
  title={Dinov3},
  author={Sim{\'e}oni, Oriane and Vo, Huy V and Seitzer, Maximilian and Baldassarre, Federico and Oquab, Maxime and Jose, Cijo and Khalidov, Vasil and Szafraniec, Marc and Yi, Seungeun and Ramamonjisoa, Micha{\"e}l and others},
  journal={arXiv preprint arXiv:2508.10104},
  year={2025}
}

@article{zhang2025qwen3embedding,
  title={Qwen3 embedding: Advancing text embedding and reranking through foundation models},
  author={Zhang, Yanzhao and Li, Mingxin and Long, Dingkun and Zhang, Xin and Lin, Huan and Yang, Baosong and Xie, Pengjun and Yang, An and Liu, Dayiheng and Lin, Junyang and others},
  journal={arXiv preprint arXiv:2506.05176},
  year={2025}
}

@article{kuhn1955Hungarian,
author = {Kuhn, H. W.},
title = {The {Hungarian} method for the assignment problem},
journal = {Naval Research Logistics Quarterly},
volume = {2},
number = {1-2},
pages = {83-97},
doi = {https://doi.org/10.1002/nav.3800020109},
url = {https://onlinelibrary.wiley.com/doi/abs/10.1002/nav.3800020109},
eprint = {https://onlinelibrary.wiley.com/doi/pdf/10.1002/nav.3800020109},
year = {1955}
}

@article{vaswani2023attentionneed,
  title={Attention is all you need},
  author={Vaswani, Ashish and Shazeer, Noam and Parmar, Niki and Uszkoreit, Jakob and Jones, Llion and Gomez, Aidan N and Kaiser, {\L}ukasz and Polosukhin, Illia},
  journal={Advances in neural information processing systems},
  volume={30},
  year={2017}
}

@INPROCEEDINGS{hadsell206,
  author={Hadsell, R. and Chopra, S. and LeCun, Y.},
  booktitle={IEEE Computer Society Conference on Computer Vision and Pattern Recognition (CVPR)}, 
  title={Dimensionality Reduction by Learning an Invariant Mapping}, 
  year={2006},
  volume={2},
  number={},
  pages={1735-1742},
  doi={10.1109/CVPR.2006.100}}

@article{oord2019representation,
  title={Representation learning with contrastive predictive coding},
  author={Oord, Aaron van den and Li, Yazhe and Vinyals, Oriol},
  journal={arXiv preprint arXiv:1807.03748},
  year={2018}
}

@inproceedings{fan2016pointsetgenerationnetwork,
  title={A point set generation network for {3D} object reconstruction from a single image},
  author={Fan, Haoqiang and Su, Hao and Guibas, Leonidas J},
  booktitle={Proceedings of the IEEE conference on computer vision and pattern recognition},
  pages={605--613},
  year={2017}
}
}

\clearpage
\appendix
\section*{Table of Contents}

\label{sec:appendix}

\startcontents[mytoc] 

\printcontents[mytoc] 
{l}{1} 
{\setcounter{tocdepth}{2}} 

\section{Detailed Performance Analysis}

\subsection{Effect of the Number of Steps}

\begin{figure}[htbp]
    \centering
    \begin{subfigure}[t]{\linewidth}
        \begin{subfigure}[t]{0.49\linewidth}
            \centering
            \includegraphics[width=\linewidth]{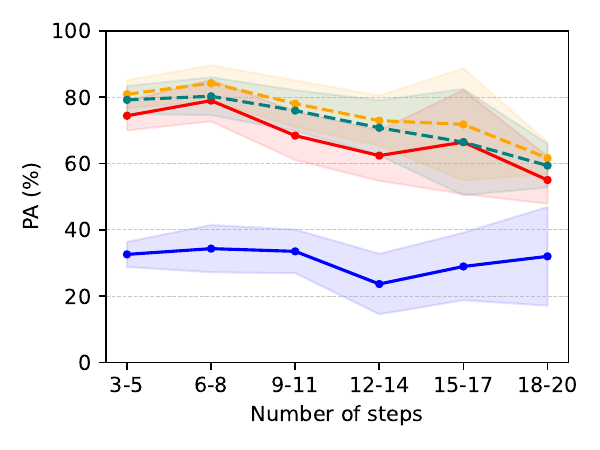}
        \end{subfigure}
        \hfill
        \begin{subfigure}[t]{0.49\linewidth}
            \centering
            \includegraphics[width=\linewidth]{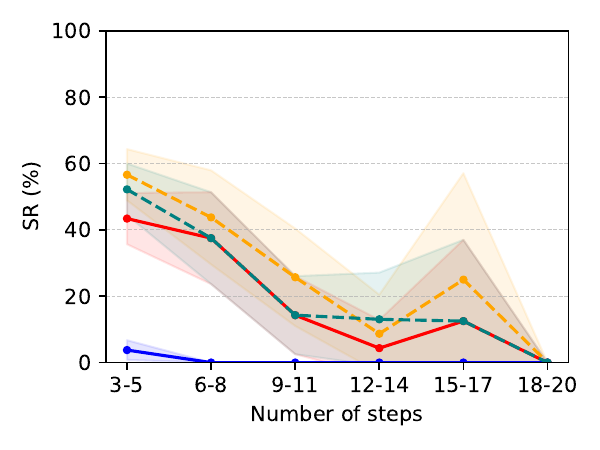}
        \end{subfigure}
        \caption{GT order setting}
    \end{subfigure}
    \begin{subfigure}[t]{\linewidth}
        \begin{subfigure}[t]{0.49\linewidth}
            \centering
            \includegraphics[width=\linewidth]{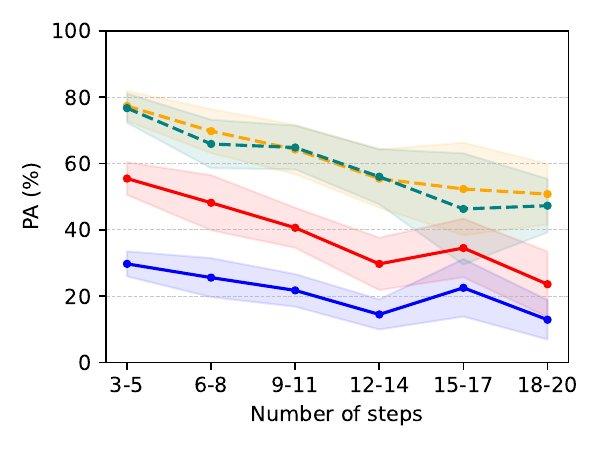}
        \end{subfigure}
        \hfill
        \begin{subfigure}[t]{0.49\linewidth}
            \centering
            \includegraphics[width=\linewidth]{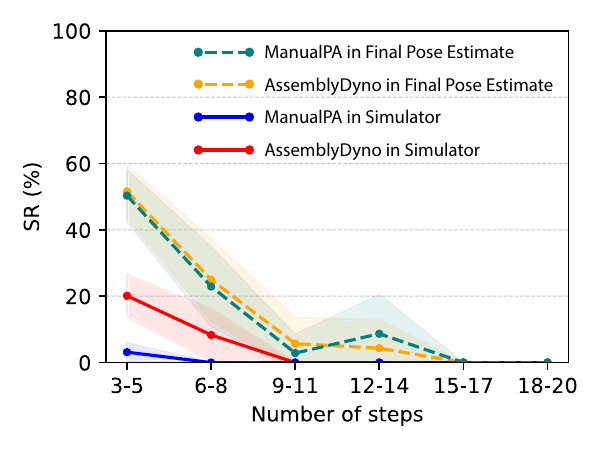}
        \end{subfigure}
        \caption{Standard setting}
    \end{subfigure}
    \caption{\textbf{Performance as a function of number of steps}. We present the PA and SR metric in two evaluation protocols (static final pose and simulation), for \modelName and ManualPA in two experiment settings. Shaded areas represent 95\% confidence intervals of the metric.}
    \label{fig:step-sim-metric}
\end{figure}

As shown in Figure~\ref{fig:step-sim-metric}, across both settings (GT order and Standard) and both metrics (PA, SR), all curves decline as the number-of-step bin increases, indicating that longer sequences, which contain more complex diagrams and assembled geometries, are harder. SR drops more steeply than PA and often approaches zero for long sequences under simulation, acting as the most strict metric in our study.

Meanwhile, it shows \modelName is more robust in simulation. In the simulation protocol (solid lines), \emph{\modelName} (red) consistently lies above \emph{ManualPA} (blue) for both PA and SR across nearly all number-of-step bins and in both settings, showing stronger execution robustness.

Again, the figure demonstrates our simulation is the stricter evaluation. For every method, metric, and setting, solid lines are lower than dashed lines (final-pose evaluation), confirming that simulation reveals failures that static end-state checks miss.

\subsection{Effect of Trajectory Category}

We compare our work against the ManualPA baseline~\cite{zhang2025manual} across multiple trajectory categories and two evaluation settings. As shown in Table~\ref{tab:category-sim-metrics}, both approaches perform well on stationary trajectories. Stationary trajectories are always the first step of the assembly, where both the context and motion are simple. In contrast, for complex motions such as rotational or insert-and-rotate trajectories, the performance of both methods drops, but our approach remains substantially more robust, achieving noticeably higher PA and lower geometric error. Overall, across most categories and in both settings, our method outperforms ManualPA, doubling the improvement in PA.

\begin{table}[htbp]
\footnotesize
\caption{
\textbf{Comparison of assembly performance across trajectory categories.}
For each category, we report three metrics computed using our simulation-based evaluation protocol: median Average Chamfer Distance (mACD), median Final Chamfer Distance (mFCD), and Percentage of Accurate assemblies (PA). We compare them against corresponding results from the baseline (\textit{ManualPA}\cite{zhang2025manual}). 
}

    \label{tab:category-sim-metrics}
    \centering
\begin{tabular*}{\linewidth}{@{\extracolsep{\fill}}lcccccc@{}}
\toprule
\textbf{Category} &
\multicolumn{2}{c}{\textbf{mACD ($10^{-3}$)$\downarrow$}} &
\multicolumn{2}{c}{\textbf{mFCD ($10^{-3}$)$\downarrow$}} &
\multicolumn{2}{c}{\textbf{PA (\%)$\uparrow$}} \\
& Ours & \cite{zhang2025manual} & Ours & \cite{zhang2025manual} & Ours & \cite{zhang2025manual}\\
\midrule
\textit{Standard Setting} \\
All            & \textbf{34.82} & 192.79 & \textbf{15.97} & 56.04 & \textbf{42.9} & 23.2 \\
Stationary     & \textbf{5.92}  & 108.21 & 5.72  & \textbf{4.41}  & 61.1 & \textbf{62.9} \\
Translational  & \textbf{33.73} & 187.06 & \textbf{15.36} & 50.92 & \textbf{43.8} & 24.5 \\
Rotational     & \textbf{50.48} & 272.77 & \textbf{26.76} & 159.05 & \textbf{30.6} & 5.6 \\
Insertion      & \textbf{147.83} & 250.08 & \textbf{10.15} & 40.26 & \textbf{48.8} & 26.7 \\
Insert + Rotate & 206.59 & \textbf{189.52} & \textbf{44.20} & 82.64 & \textbf{14.3} & 0.0 \\
\midrule
\textit{GT Order Setting} \\
All            & \textbf{9.97} & 230.52 & \textbf{4.43} & 35.00 & \textbf{70.1} & 31.3 \\
Stationary     & \textbf{2.78} & 120.48 & 2.72 & \textbf{2.53}  & \textbf{77.5} & 76.4 \\
Translational  & \textbf{9.43} & 224.32 & \textbf{4.30} & 30.25 & \textbf{70.9} & 33.1 \\
Rotational     & \textbf{14.49} & 373.30 & \textbf{7.21} & 215.70 & \textbf{59.7} & 6.5 \\
Insertion      & \textbf{100.24} & 288.18 & \textbf{5.92} & 40.24 & \textbf{73.3} & 34.9 \\
Insert + Rotate & \textbf{117.38} & 166.57 & \textbf{10.36} & 67.83 & \textbf{42.9} & 0.0 \\
\bottomrule
    \end{tabular*}
\end{table}

\subsection{Effect of Loss Design}

\paragraph{Ablation Study.}
From the ablation results in Table~\ref{tab:ablation1}, we observe that all loss components in our framework are essential. 
Removing any individual loss term (\(\mathcal{L}_{P}\), \(\mathcal{L}_{T}\), \(\mathcal{L}_{R}\), or \(\mathcal{L}_{SR}\)) consistently degrades performance across both the \emph{Final Pose Estimate} metrics and the \emph{Assembly in Simulator} metrics. 
The drops are particularly notable in the simulation-based metrics (mACD, mFCD, PA, and SR), indicating that each loss contributes critically to enabling the model to produce physically executable assembly trajectories. 
These observations confirm the necessity of the full loss design used in \modelName.

\paragraph{Sensitivity Analysis}
We further conduct a sensitivity analysis by varying the weights of the rotational loss \(\lambda_{R}\) and the rotation-regularization loss \(\lambda_{SR}\). The original weights are in Table~\ref{tab:training-hparams}. 
Across the tested weight settings, the resulting performance metrics exhibit only small fluctuations. 
This low variance indicates that our method is robust to moderate perturbations of these hyperparameters. 
Thus, within the examined range, the overall assembly performance is not highly sensitive to the specific weight choices of these losses, demonstrating stability of the training objective.

\begin{table*}[htbp]
  \footnotesize
  \caption{\textbf{Part assembly results on the test split of \datasetName}. 
  Best results are in \textbf{bold}, second best in \textcolor{blue}{blue}.}
  \centering
  \begin{tabular*}{\linewidth}{@{\extracolsep{\fill}}lccccccc@{}}
    \toprule
    \textbf{Model} 
    & \multicolumn{3}{c}{\textbf{Final Pose Estimate}} 
    & \multicolumn{4}{c}{\textbf{Assembly in Simulator}} \\
    \cmidrule(lr){2-4} \cmidrule(lr){5-8}
    & SCD($10^{-3}$)$\downarrow$
    & PA(\%)$\uparrow$
    & SR(\%)$\uparrow$
    & mACD($10^{-3}$)$\downarrow$
    & mFCD($10^{-3}$)$\downarrow$
    & PA(\%)$\uparrow$
    & SR(\%)$\uparrow$ \\
    \midrule

    \textbf{\modelName}
      & 3.87 
      & \textbf{79.69}
      & 44.29
      & \textcolor{blue}{9.97}
      & \textcolor{blue}{4.43}
      & \textcolor{blue}{70.15}
      & \textbf{33.57} \\

    \quad w/o $\mathcal{L}_P$
      & 4.09 & 77.75 & 38.57
      & 10.79 & 5.11 & 67.91 & 27.50 \\

    \quad w/o $\mathcal{L}_T$
      & 4.19 & 68.43 & 27.86
      & 17.69 & 6.92 & 57.41 & 25.00 \\

    \quad w/o $\mathcal{L}_R$
      & 3.97 & 73.71 & 37.14
      & 11.94 & 5.26 & 64.57 & 26.43 \\

    \quad w/o $\mathcal{L}_{S_R}$
      & 3.85 & 79.17 & 43.21
      & 16.98 & 11.06 & 47.18 & 5.36 \\

    \midrule
    \textit{Sensitivity Analysis}\\

    \quad $\lambda_R = 1$
      & 3.88 & 78.64 & 42.86
      & 10.15 & 4.59 & 69.11 & 30.00 \\

    \quad $\lambda_R = 10$
      & \textcolor{blue}{3.74} & 79.49 & \textbf{46.43}
      & \textbf{9.16} & \textbf{4.24} & \textbf{70.20} & \textcolor{blue}{32.14} \\

    \quad $\lambda_{S_R} = 1$
      & \textbf{3.71} & \textcolor{blue}{79.61} & \textcolor{blue}{44.64}
      & 10.93 & 5.65 & 64.35 & 20.00 \\

    \quad $\lambda_{S_R} = 10$
      & 3.82 & 79.20 & 40.71
      & 10.04 & \textbf{4.24} & 70.04 & 30.36 \\
    \bottomrule
  \end{tabular*}
  \label{tab:ablation1}
\end{table*}

\subsection{Effect of Text Instructions}
\label{sec:text-effect}
\begin{figure}[htbp]
    \centering
    \begin{subfigure}[b]{0.55\linewidth}
        \centering
        \includegraphics[width=\linewidth]{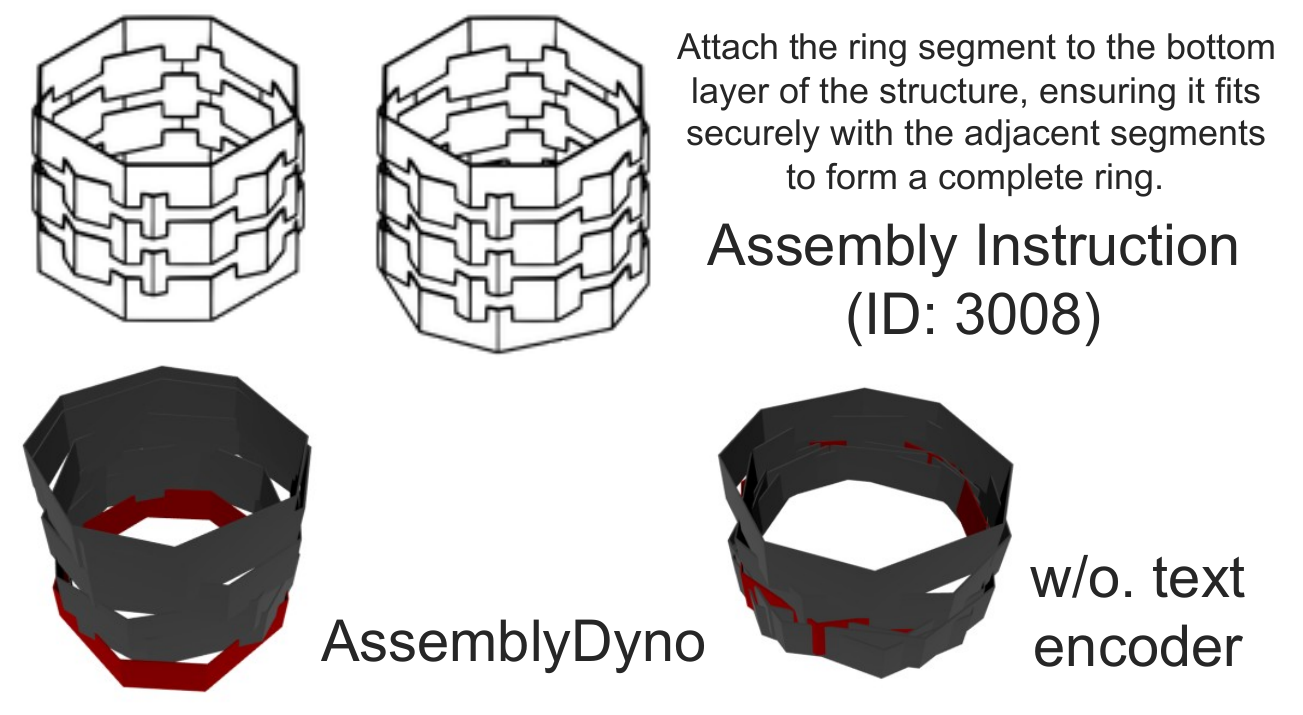}
        \caption{}
        \label{fig:text-example}
    \end{subfigure}
    \begin{subfigure}[b]{0.4\linewidth}
        \centering
        \includegraphics[width=\linewidth]{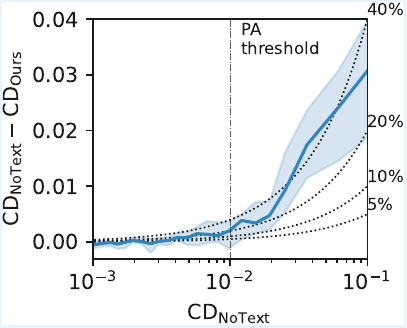}
        \caption{}
        \label{fig:cd}
    \end{subfigure}
    \caption{Text helps for difficult cases. \textbf{(a)} Example instruction step (top) and resulting prediction with vs. without using text (bottom). \textbf{(b)} Median value of 
    $\mathrm{CD}_{\text{NoText}} - \mathrm{CD}_{\text{Ours}}$, plotted vs. $\mathrm{CD}_{\text{NoText}}$. 
    }
    \label{fig:combined_metrics}
\end{figure}

While the aggregated performance gains in Table 2 may appear marginal (+2\%), the influence of text is not sufficiently unveiled. Specifically, for challenging assemblies such as the one in Fig.~\ref{fig:text-example}, we find that incorporation of text leads to significant benefits. 
In Fig.~\ref{fig:cd}, we plot the median improvement in part-wise chamfer distance (CD) by \modelName, as a function of CD of the text-omitted version. Clearly, text instructions yield significant gains 
(up to 40\%) for the worst cases ($\mathrm{CD}>10^{-2}$, the right half of the plot).
This is currently not reflected in the PA metric we report, as it only counts the fraction of parts with $\mathrm{CD}<10^{-2}$. We will include this in the final paper.

\subsection{Adding Multiple Parts in One Step} 
To evaluate the robustness of our model when multiple parts are added in a single step, we trained and tested our model while randomly removing (masking) the diagrams for up to 2 steps from each assembly. Table~\ref{tab:maskout} shows that our model is significantly more robust to missing step diagrams than the Manual-PA baseline.  

\begin{table}[htbp]
    \footnotesize
    \caption{When step diagrams are randomly masked out (keeping text), AssemblyDyno is much more robust than ManualPA.}
    \vspace{-6px}
    \centering
    \setlength{\tabcolsep}{4pt}
    \setlength{\aboverulesep}{0pt}

    \begin{tabularx}{\columnwidth}{l c c c c} 
        \toprule
        \textbf{Model} 
        & \textbf{Masked?} 
        & \textbf{SCD($10^{-3}$)$\downarrow$} 
        & \textbf{PA(\%)$\uparrow$} 
        & \textbf{SR(\%)$\uparrow$} \\
        \midrule

        \multirow{2}{*}{\textbf{\modelName}} & -  
            & 3.81 & 77.08 & 40.36 \\ 

        & \checkmark 
            & \textbf{4.61} & \textbf{69.20} & \textbf{22.86} \\

        \midrule

        \multirow{2}{*}{\textbf{ManualPA}} & -
            & 4.15 & 77.40 & 39.28 \\
        & \checkmark
            & 6.02 & 60.90 & 9.286 \\

        \bottomrule
    \end{tabularx}
    \label{tab:maskout}
\end{table}

\section{Physics-aware or physics-in-the-loop?}

In this paper, we use the term ``physics-aware'' to indicate that a physics engine is used to generate the ground-truth part trajectories in our dataset and to evaluate the assemblies predicted by models. 

We attempted a ``physics-in-the-loop'' strategy for our model, where physics constraint signals are included in the training loss.  
However, we found it to be less effective than supervised training using ground-truth assembly trajectories. 

\begin{figure}[htbp]
    \centering
\includegraphics[width=\linewidth]{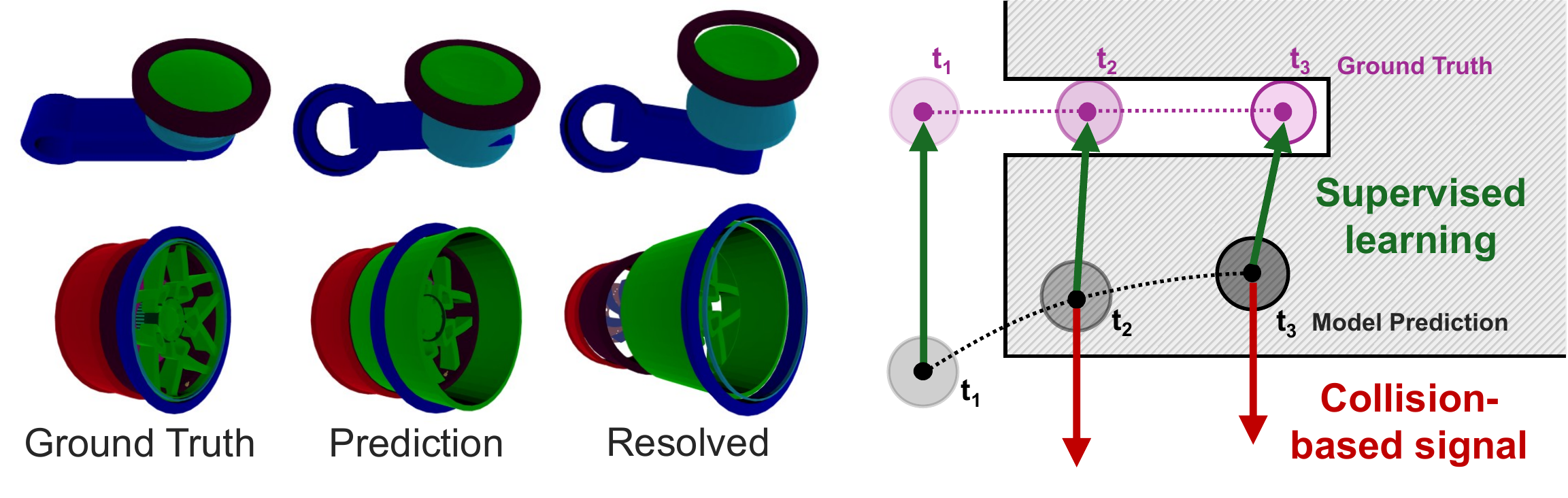}
    \caption{\textbf{Left}: Physics-based post-processing of an assembly. \textbf{Right}: Supervised learning  guides predictions towards ground truth, while collision forces push parts to the nearest free space.}
    \label{fig:sdf}
\end{figure}

This claim is supported by experiments where we use simulator refinements at test time: when predicted final part poses exhibit even minor interpenetrations (on the order of mm), we use a physics simulator to resolve these by pushing parts apart, resulting in large displacements from the ground-truth poses (see Figure~\ref{fig:sdf} left). Such refinements is detrimental, reducing PA from 79.69\% to 70.53\% and SR from 44.29\% to 36.79\%. 

In Figure~\ref{fig:sdf} right, we conceptually illustrate this difficulty using a hypothetical ground-truth part trajectory and the directions of collision-induced forces from the simulator. As shown, when ground-truth trajectories are available, they can provide substantially stronger and more stable learning signals than the noisy collision gradients from the simulator. Consequently, designing effective collision-aware losses or post-processing schemes would require significant innovations beyond the scope of this work.

\section{Physics Simulator Configurations}

\begin{figure}[htbp]
    \centering
    \includegraphics[width=\linewidth]{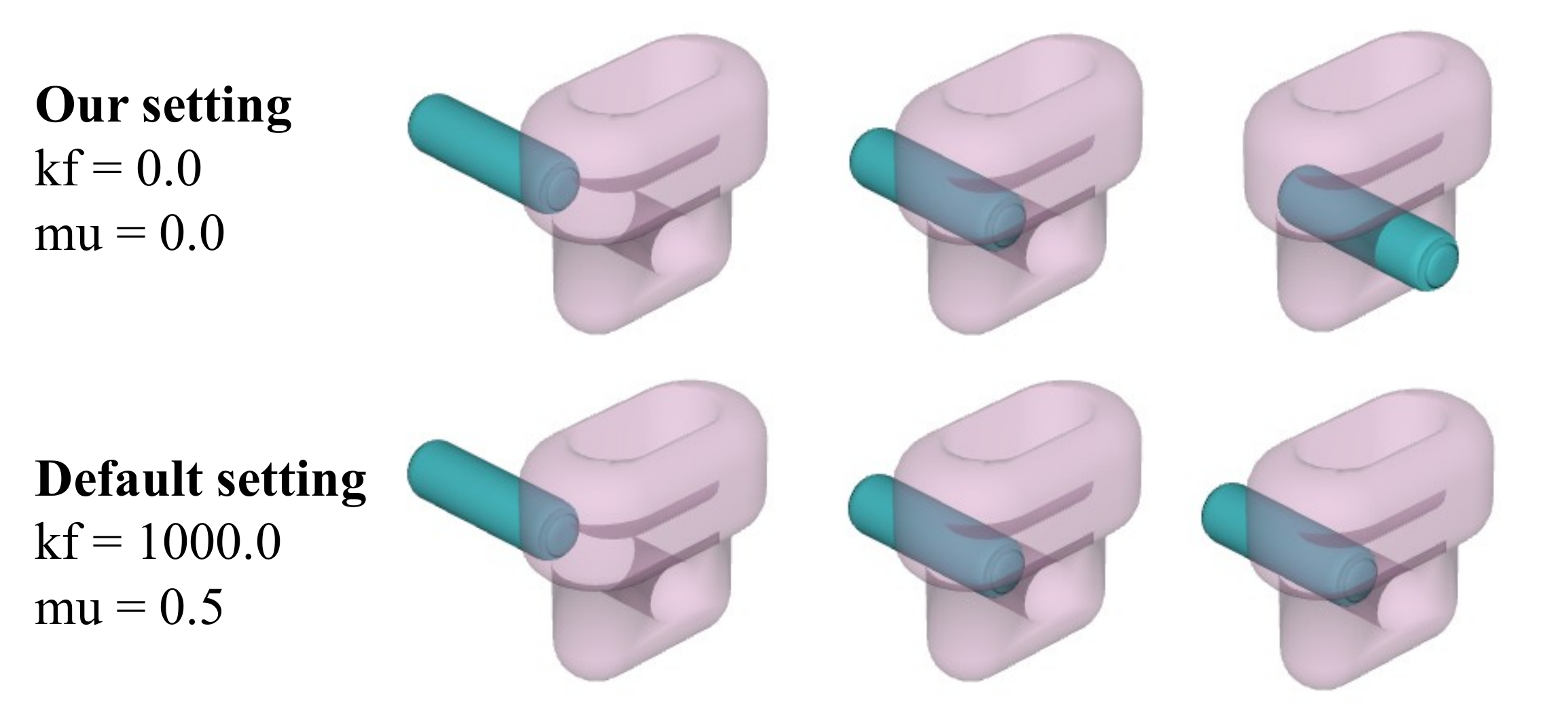}
    \caption{\textbf{Effects of Friction Parameters}. (Top) when disabling the friction effects in the simulator (our setting), the orange part can be successfully installed under the guidance of ground truth trajectory. (Bottom) default friction setting leads to a stuck at the rim of the hole.}
    \label{fig:friction-param}
\end{figure}

We present the most important simulator configurations in Table~\ref{tab:sim-settings}, which consists of general settings such as gravity and simulation substeps, as well as material settings that determined the friction behaviors of the shape. The friction parameters \texttt{kf} and \texttt{mu} need to be set to $0$ in our evaluation. as shown in Figure~\ref{fig:friction-param}, when friction effects are not disabled, some insertion behaviors will not be executed, even if we use ground truth trajectories to guide the simulator.

While we use a specific simulator~\cite{newton2025} in our study, our evaluation protocol (\emph{i.e.} the simulation design and its metrics) is agnostic to simulator choice, as long as it supports collision detections on non-convex mesh geometries.

\begin{table}[htbp]
\footnotesize
\caption{Hyperparameters for training \modelName.}
\label{tab:training-hparams}
\centering
\begin{tabular*}{\linewidth}{@{\extracolsep{\fill}}llll@{}}
\toprule
\textbf{Name} & \textbf{Value} & \textbf{Name} & \textbf{Value} \\
\midrule
Batch size & 64 & Epoch & 1,000 \\
Optimizer & AdamW & Learning rate & $4\times10^{-5}$ \\
Weight decay & $1\times10^{-4}$ & Betas for AdamW & $(0.9, 0.999)$ \\
$\lambda_{P}$ & 20 & $\lambda_{T}$ & 1 \\
$\lambda_{R}$ & 20 & $\lambda_{S_T}$ & 1 \\
$\lambda_{S_R}$ & 20 & & \\
\bottomrule
\end{tabular*}
\end{table}

\begin{table}[htbp]
\footnotesize
    \caption{\textbf{Simulator Configurations}. The most important parameters contain general settings (first two parameters) and material settings (the remaining). The parameters that differ from default simulator settings are \textbf{bolded}. }
    \label{tab:sim-settings}
    \centering
    \begin{tabular*}{\linewidth}{@{\extracolsep{\fill}}llc@{}}
    \toprule
       \textbf{Parameter} & \textbf{Description} & \textbf{Value}\\
    \midrule
       \texttt{gravity} & The gravity force. & \textbf{0.0}\\
       \texttt{substeps} & Substeps between trajectory waypoints. & 60\\
       \texttt{ka} & The contact adhesion distance. & 0.0\\
       \texttt{kd} & The contact damping stiffness. & 1000.0\\
       \texttt{ke} & The contact elastic stiffness. & 100000.0\\
       \texttt{kf} & The contact friction stiffness. & \textbf{0.0}\\
       \texttt{mu} & The coefficient of friction. & \textbf{0.0}\\
       \texttt{restitution} & The coefficient of restitution. & 0.0\\
       \texttt{thickness} & The thickness of the shape. & 1e-05\\
    \bottomrule
    \end{tabular*}
\end{table}

\section{Performance of Classic Motion Planning}

We choose RRT and RRT-connect (used in \cite{tie2025manual2skill}) to demonstrate classical motion planning methods are not suitable for the assembly trajectory generations in our scenario where the final poses are predicted. We conduct this experiment on a Windows Subsystem for Linux with an Intel Core i9-14900K CPU (32 cores) and 128 GB RAM.

In the following experiments, we feed the predicted final poses from \modelName to the two motion planning methods, instructing them to calculate the corresponding assembly motion. We inspect if they can provide solutions, no matter the quality, within given time constraints.

Table~\ref{tab:sr-classic} highlights two major limitations of classical motion planners.
First, these methods are inherently slow and struggle to find solutions under practical
time constraints. Even with generous limits such as 30\,s or 60\,s, both RRT and
RRT-Connect achieve success rates below 10\%, indicating that they rarely return
solutions quickly enough to be useful in real-world assembly scenarios.

More importantly, even when a very long time limit is imposed (120s), classical planners still fail to solve more than a small fraction of the tasks. This shows that classical planners require strictly collision-free goal states. However, the predicted final poses from \modelName may contain minor shape overlaps or small interpenetrations between parts, which are unavoidable when predictions are generated by learning-based models. These small inconsistencies cause classical planners to reject the goal configuration or become stuck while attempting to resolve infeasible collisions, preventing them from producing valid trajectories even with unlimited
compute.

Assemble-them-all uses a search heuristic to produce part trajectories, whose computational complexity is combinatorial in the number of parts. While it takes 6.7s to produce assemblies on average, we found that tail cases take much longer, e.g., only 57.5\% achieve success at $\leq30s$. Instead, our \modelName predicts all part trajectories \textit{in a single forward pass}. 

\begin{table}[]
\caption{\textbf{Success rate of non-neural motion planning}. We present the success rate of returning answers (regardless of their quality) under varying time constraints. We use the predicted final poses from \modelName as inputs.}
\label{tab:sr-classic}
\footnotesize
\centering
\begin{tabular*}{\linewidth}{@{\extracolsep{\fill}}lccccccc@{}}
\toprule
\textbf{Algorithm} & \multicolumn{7}{c}{\textbf{Success Rate (\%) with Timeout Constraints}} \\
\cmidrule(lr){2-8}
 & 1s & 2s & 5s & 10s & 30s & 60s& 120s\\
\midrule
RRT & 2.9&3.6&4.6&5.0&6.4&6.8& 7.1 \\
RRT-Connect & 2.9&3.2&3.6&4.6&5.4&6.1& 7.1 \\
Assemble Them All & 4.6 & 8.6 & 21.8 & 34.6 & 57.5 & 72.1 & 81.8\\
\bottomrule
\end{tabular*}
\end{table}

\section{Qualitative Results}

\begin{figure*}[htbp]
    \centering
\includegraphics[width=\linewidth]{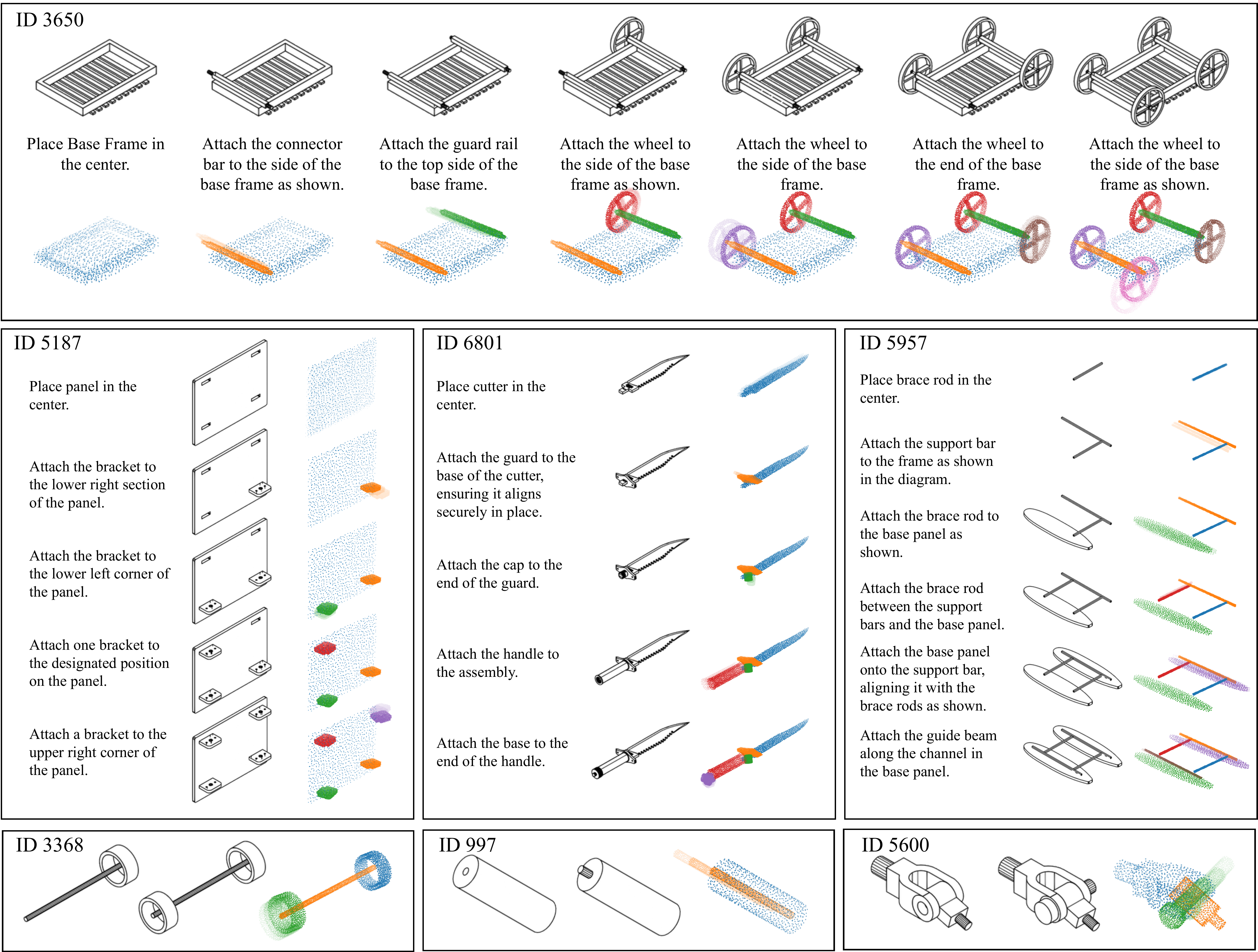}
    \caption{\textbf{User manuals with predicted trajectories}. We present the user manual and predicted trajectories of \modelName as the colored point clouds. Top two rows illustrate complete user manuals while the last row features insertion assembly steps. Multiple time steps are overlapped as transparent layers. The trajectories are executed in the stimulator, showing their outcomes when considering physical constraints.}
    \label{fig:more-examples}
\end{figure*}

Figure~\ref{fig:more-examples} illustrates user-manual assembly instructions alongside our model’s predicted trajectories. For each step, we visualize the motion as a sequence of temporally ordered point-cloud snapshots rendered as semi-transparent overlays. Although each trajectory contains 12 time steps, we display only the 1st, 6th, and 12th steps to provide a clear yet concise depiction of the object’s motion during assembly. These overlaid time-step frames highlight how the predicted trajectories evolve over time and how the parts move toward their final configurations in each manual step.

\section{Limitation}
Unlike IKEA-style furniture parts, which often have significant part symmetries and duplications, the complex industrial parts in \datasetName
(including large variation in part sizes) 
are found to be sensitive to even minor changes in the camera viewpoints. We believe new approaches for view-invariant diagram representations are necessary, and our multiview data generation pipeline facilitates research into this important topic.

\section{Dataset Construction Details}

\subsection{User Study of the User Manuals}
\begin{figure}[htbp]
    \centering
    \includegraphics[width=\linewidth]{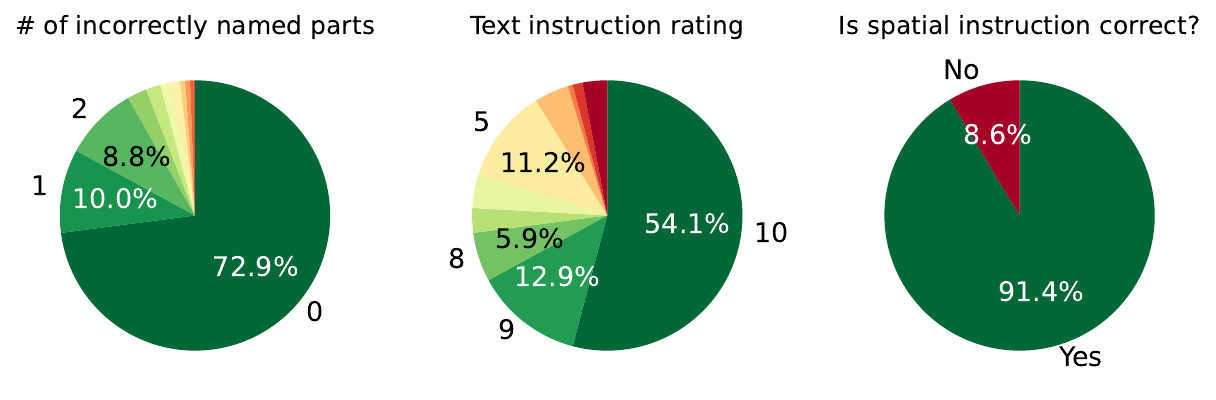}
    \caption{\textbf{User study on text annotation quality}. We present: (left) the distribution of incorrect part name amounts within a sampled assembly shape; (middle) the distribution of subjective text instruction ratings among all sampled shapes, higher is better; (right) the proportion of correct spatial information in the text instructions.}
    \label{fig:survey-results}
\end{figure}

We conduct a two-stage user survey to evaluate the quality of our part names and text instructions. First, participants are shown 200 assembled CAD shapes, each with two rendered views and color-labeled parts, and are asked to count incorrectly named parts (\emph{e.g.}, a cube labeled as ``sphere''). As shown in Fig.~\ref{fig:survey-results}, 72.9\% of shapes contain zero incorrect names, and over 90\% contain at most one or two, indicating high semantic accuracy.

Second, for each assembly we sample one instruction step and provide the current diagram, prior diagram, text instruction, and a labeled reference image. Participants rate text quality (1--10) and verify whether the spatial instruction matches the diagram. Results show that 54.1\% of text instructions receive a rating of 10, and over two-thirds score 9 or above. Spatial correctness is also high: 91.4\% of instructions are labeled correct. These outcomes confirm that our part names, textual descriptions, and spatial references are reliably annotated, providing a strong foundation for downstream tasks.

\subsection{Choose Diagram Camera Views}
As we render all diagrams in a set of camera views, for each shape assembly, We apply a heuristic to select the camera view that best demonstrate the assembly process. 
The objective is to choose, for each assembly part, a camera view that provides strong visual coverage of the part during the relevant assembly step and in the final assembled state. The method proceeds in three conceptual stages: (1) measuring visibility, (2) scoring and normalizing per-part visibility, and (3) combining scores across the entire assembly to produce a consistent camera assignment.

\paragraph{Visibility Measurement}

Consider a set of cameras indexed by \(c \in \mathcal{C}\), and a sequence of assembly parts indexed in order by \(p \in \mathcal{P}\). For each camera \(c\) and part \(p\), we observe two images:

\begin{itemize}
    \item the diagram taken at the assembly step when part \(p\) is assembled, and
    \item the diagram taken at the final assembly completion.
\end{itemize}

From each diagram image we extract the number of pixels belonging to part \(p\). Let
$n_{c,p}^{(A)} \quad \text{and} \quad n_{c,p}^{(F)}$
denote, respectively, the number of pixels of part \(p\) visible in camera \(c\) during the assembly step and during the final step.
Thus, visibility is characterized by the pair
$\bigl(n_{c,p}^{(A)},\; n_{c,p}^{(F)}\bigr).$

\paragraph{Per-Part Visibility Score}

Visibility should contribute to the score in a way that has diminishing returns for large pixel counts. A logarithmic visibility score fulfills this requirement. For each camera \(c\) and part \(p\), define:

\[
s_{c,p} =
\begin{cases}
\log\!\left(1 + \lambda\, n_{c,p}^{(A)}\right)
+
\log\!\left(1 + \lambda\, n_{c,p}^{(F)}\right),
& \text{if } n_{c,p}^{(A)} > 0, \\[8pt]
0, & \text{if } n_{c,p}^{(A)} = 0,
\end{cases}
\]

where $\lambda$ is a scaling constant set as $0.05$ that moderates the influence of raw pixel counts.

This construction ensures:
\begin{itemize}
    \item both assembly-step visibility and final-step visibility contribute additively,
    \item visibility in the assembly step is essential (otherwise the score is zero),
    \item visibility contributions grow sublinearly.
\end{itemize}

For each part \(p\), the visibility scores \(\{s_{c,p}\}_{c \in \mathcal{C}}\) are divided by the max score across all cameras ($\max_{c \in \mathcal{C}} s_{c,p}$), so that the best camera attains a normalized score $\hat{s}_{c,p}$.

\paragraph{Aggregated Camera Quality Across All Parts}

To determine which cameras are most useful across the entire assembly process, the normalized per-part scores are summed over all parts to \(S_c\).

\[
S_c = \sum_{p \in \mathcal{P}} \hat{s}_{c,p}.
\]

The quantity \(S_c\) expresses the overall usefulness of camera \(c\) across the entire assembly. The cameras are ranked in descending order of \(S_c\).
This global ranking reflects which cameras tend to provide good visibility for many parts.

\subsection{Instructional Text Generation}

We use VLMs to name the CAD parts one by one. For each CAD part, we provide the VLM with the following text prompts and three images:
\begin{itemize}
    \item The diagram from the first step where the part is introduced, with the part highlighted.
    \item The final-step diagram of the completed assembly, also highlighting the same part.
    \item When there are similar parts in the assembly, we provide an additional diagram where all its counterparts are highlighted.
\end{itemize} 

\begin{lstlisting}[language=Markdown]
You are a product design assistant who write user manuals.
You should name the colored component in the first image.
In the second image, you are given the final assembled model with your focus component colored.
In the third image, all similar components are colored for your reference. You should give a general name (in singular form) for all these similar components.

- Adopt one name. Don't include multiple choices.
- Don't include color into names.
- Avoid using oriental words such as left, right, horizontal.
- Avoid existing part names: {existing_names}

Think in steps and finalize your answer in json.
Example output:
```json
{
  "name": "Feet"
}
```
```json
{
  "name": "Connector Bar"
}
```
\end{lstlisting}

We find similar parts by grouping their bounding box sizes. As we prompt the VLM to assign a general name for each group, we only name one part within the group.
 
For all diagrams, we choose the best camera view from a predefined set, by colorizing the target part and selecting the view that maximizes the number of colored pixels of our target part in the diagram image. The camera views vary among CAD parts, which is different from final user manuals, where all diagrams share the same view within the shape assembly.

We use the generated names to create text instructions for final user manuals. We generate assembly text instructions step-by-step. Within each VLM call, we provide the following text prompts with two images, where the camera views align with the final user manuals.:
\begin{itemize}
    \item The diagram of the current assembly step, with the part to be assembled highlighted in color.
    \item The same diagram of the same step, but with all of the parts highlighted in varying colors and labeled by their names. 
\end{itemize}

\begin{lstlisting}[language=Markdown]
You are a product design assistant who write user manuals.
You should describe the assembly step of the first image, which involves only one component highlighted in color.
The names of the current component and previous components are shown in the second image as a reference.

- Just describe the current assembly step.
- Don't include instructions for previous steps and components.
- Correct the incorrect plural form of the component name if any.
- Don't subdivide the current step into multiple sub-steps.
- Avoid words like "as shown in the image" or "as illustrated".
- Don't include color in your description.


Think in steps and finalize your answer in json.
Example output:
```json
{
  "text": "your description here"
}
```
\end{lstlisting}

\end{document}